\documentclass[11pt]{article}

\usepackage[final]{acl}

\usepackage{times}
\usepackage{latexsym}

\usepackage[T1]{fontenc}

\usepackage[utf8]{inputenc}

\usepackage{microtype}

\usepackage{inconsolata}

\usepackage{enumitem}
\usepackage{amsmath}
\usepackage{amssymb} 
\usepackage{graphicx}
\usepackage{booktabs}
\usepackage{hyperref}
\usepackage{url}
\usepackage{array}  
\usepackage[table]{xcolor}
\usepackage{relsize}
\usepackage{multirow}
\usepackage{pifont}
\usepackage{wrapfig}
\usepackage{subcaption}
\usepackage{diagbox}

\title{Causal2Vec: Improving Decoder-only LLMs as Embedding Models through a Contextual Token}

\author{
    Ailiang Lin$^{1}$\thanks{$\,$ Equal Contribution.}, 
    Zhuoyun Li$^{2}$\footnotemark[1], 
    Yusong Wang$^{1}$,
    Kotaro Funakoshi$^{1}$,
    Manabu Okumura$^{1}$\\
    $^{1}$Institute of Science Tokyo \quad $^{2}$Tencent \\
\texttt{\{linailiang, wangyi, funakoshi, oku\}@lr.first.iir.isct.ac.jp}\\ \texttt{earyli@tencent.com}
}

\begin{document}
\maketitle
\begin{abstract}
Decoder-only large language models (LLMs) have been increasingly adopted to build embedding models for diverse tasks. To overcome the inherent limitations of causal attention in representation learning, many existing methods modify the attention mechanism to be bidirectional, potentially undermining LLMs' ability to extract semantic information acquired during pre-training. Meanwhile, leading unidirectional approaches often rely on extra input text to generate contextualized embeddings, inevitably increasing computational costs. In this work, we propose Causal2Vec, a general-purpose embedding model tailored to enhance the performance of decoder-only LLMs without altering their original architectures or introducing significant computational overhead. Specifically, we first employ a lightweight BERT-style model to pre-encode the input text into a single Contextual token, which is then prepended to the LLM's input sequence, allowing each token to capture contextualized information even without attending to future tokens. Furthermore, to mitigate the recency bias introduced by last-token pooling, we concatenate the last hidden states of Contextual and EOS tokens as the final text embedding. In practice, Causal2Vec achieves a new state-of-the-art performance on the MTEB benchmark among models trained solely on publicly available retrieval datasets.
\end{abstract}

\begin{figure*}[t]
\begin{center}
\includegraphics[width=1\linewidth]{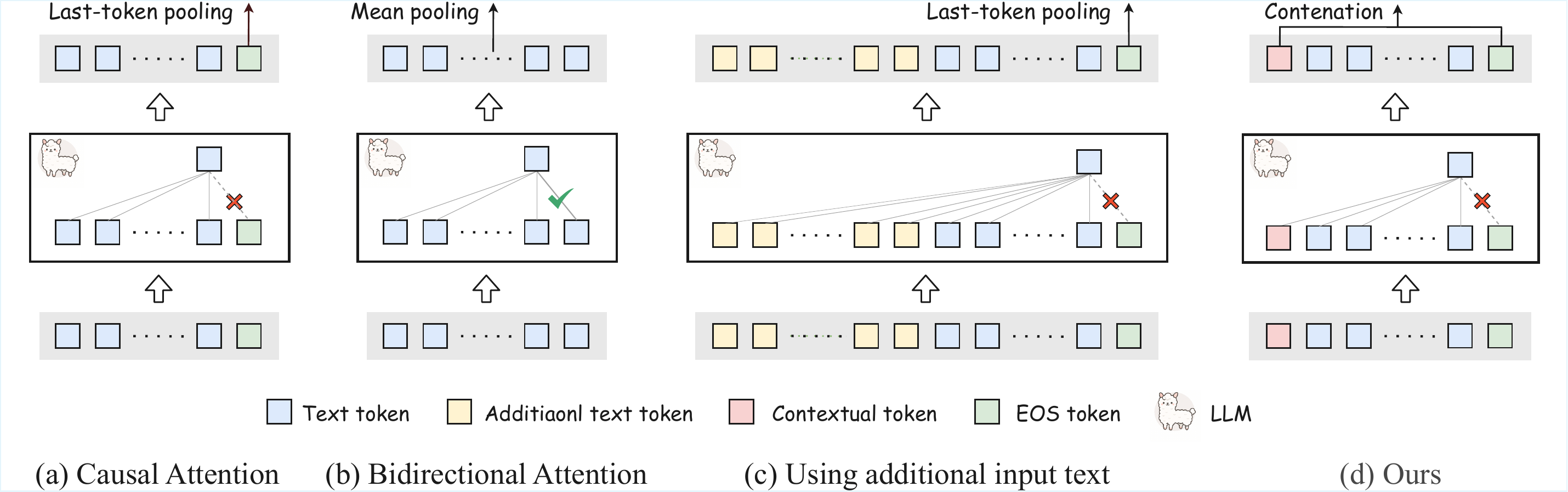}
\end{center}
\caption{Comparison of different methods to overcome the representational bottleneck of decoder-only LLMs (Fig.a). Unlike existing methods that either modify attentions from causal to bidirectional (Fig.b) or utilize extra input text (Fig.c), we prepend a \textit{Contextual token} to the LLM's input sequence (Fig.d). This allows each token to access contextualized information without attending to future tokens.}
\label{fig:fig0}
\end{figure*}

\section{Introduction}
Text embedding models encode natural language text into dense vector representations that capture contextual semantic information~\cite{text2}, enabling a wide range of downstream natural language processing (NLP) tasks, such as information retrieval, semantic textual similarity, and question answering~\citep{xiong2021approximate,mteb,cmedteb}. Moreover, embedding-based retrievers play a crucial role in enhancing the capabilities of large language model (LLM)-based Retrieval-Augmented Generation (RAG) systems~\citep{liu2024chatqa,compselect,zhang2026stable}. For many years, pre-trained language models based on encoder-only or encoder-decoder Transformer architectures, such as BERT~\citep{bert}, RoBERTa~\citep{roberta}, and T5~\citep{t5}, have been the dominant paradigm for building text embedding models~\cite{sbert,simcse,st5,e5}.

With recent advances in LLMs, considerable efforts have focused on transforming decoder-only architectures into text embedding models. However, the use of causal attention in Transformer decoders leads to incomplete information encoding for each token except the last one (Figure~\ref{fig:fig0}-(a)), significantly limiting the model's representational capacity. To address this issue, many LLM-based text embedding methods~\citep{grit,llm2vec,nvembed,mgh} achieve bidirectional attention by removing the causal attention mask, enabling each token to access the entire sequence and thus generating rich contextualized representations, as illustrated in Figure~\ref{fig:fig0}-(b). Despite notable progress, our findings in Figure~\ref{fig:causal_or_bi} highlight that modifying the original attention mechanisms of LLMs may be suboptimal for embedding tasks, as it leads to a pre-train/fine-tune attention mismatch, potentially compromising the model's ability to extract semantic information acquired during pre-training. 

In contrast, most leading causal attention-based methods~\citep{echo,bgeicl} compensate for missing contextual information in the original sequence by introducing additional input text (Figure~\ref{fig:fig0}-(c)), which inevitably increases computational costs. Moreover, to derive text embeddings from the output hidden states of LLMs, there are two mainstream strategies: mean pooling and last-token pooling. Prior studies~\citep{zhang2023language,bgeicl} show that, for embedding models based on causal attention, (weighted) mean pooling is less effective than last-token pooling, since autoregressive modeling prevents earlier tokens from attending to future tokens, leading to biased aggregated representations. As a result, last-token pooling has been more commonly adopted in unidirectional models. However, since last-token pooling typically relies on the final hidden state of the end-of-sequence (EOS) token, it can be highly sensitive to noisy information near the end of input and thus prone to recency bias~\citep{echo,nvembed}, hindering the model's ability to learn robust representations.

In this work, we propose \textbf{Causal2Vec}, a simple yet powerful causal attention-based embedding model that significantly enhances the text encoding capabilities of decoder-only LLMs, while circumventing the need to modify their original architectures or introducing significant computational overhead. Specifically, to address the representational bottleneck inherent in the causal attention mechanism while preserving the LLMs' ability to extract semantic information learned during pre-training, we first employ a lightweight, off-the-shelf bidirectional encoder to distill the contextual content of the input text into a single \textit{Contextual token}, which is then aligned to the dimensionality of LLM's word embedding space via a trainable MLP layer. As shown in Figure~\ref{fig:fig0}-(d), by prepending this token to LLM's input sequence, we enable each token to access contextualized information even under the constraints of causal masks, without switching to bidirectional attention or utilizing extra input text. Moreover, we concatenate the last hidden states of Contextual and EOS tokens as the final text embedding, effectively mitigating the recency bias introduced by last-token pooling and encouraging LLMs to better leverage the contextualized information encoded in the Contextual token.

We conduct comprehensive experiments on the MTEB benchmark~\citep{mteb} by integrating Causal2Vec into four decoder-only LLMs with parameter sizes ranging from 1.3B to 7B (S-LLaMA-1.3B, Qwen2.5-1.5B, LLaMA-2-7B, and Mistral-7B). Evaluation across 56 datasets spanning 7 tasks demonstrates that our method achieves new state-of-the-art results among models trained solely on publicly available retrieval data. Furthermore, we present extensive ablations and analyses to validate the effectiveness and necessity of the proposed mechanism. Overall, this work highlights the inherent potential of LLMs' original causal attention in generating high-quality contextualized text embeddings. The main contributions of this work are summarized below: 
\begin{itemize}
\item We introduce Causal2Vec, a simple yet powerful approach that enhances the representational capacity of LLMs without converting to bidirectional attention or requiring extra input.

\item To mitigate the representational bottlenecks of causal attention, we introduce the \textit{Contextual token} and prepend it to LLM's input sequence, allowing each token to access contextual information without attending to future ones.

\item To alleviate the recency bias introduced by last-token pooling, we concatenate the last hidden states of Contextual and EOS tokens as the final text embedding.

\item Causal2Vec achieves state-of-the-art results on MTEB among models trained solely on publicly available retrieval datasets.
\end{itemize}

\section{Related Work}
\paragraph{Bidirectional Text Embedding Models.} Over the past few years, embedding methods based on pre-trained language models with bidirectional attention, such as BERT~\citep{bert}, RoBERTa~\citep{roberta}, and T5~\citep{t5}, have dominated text embedding tasks. Early notable approaches, including SimCSE~\citep{simcse} and Sentence-T5~\citep{st5}, are pre-trained with a masked language modeling objective and fine-tuned in a contrastive manner with natural language inference (NLI) datasets. Later work like E5~\citep{e5} and GTE~\citep{gte} further improves embedding performance through weakly supervised contrastive training on curated text pair datasets. More recent methods~\citep{instrut1,instrut2,instrut3} have shifted toward developing general-purpose embedding models through task instructions, demonstrating strong generalization to unseen tasks.

\paragraph{Decoder-only LLM-based Text Embedding Models.}~\citet{luo-etal-2024-large} show that larger models with extensive pre-training consistently improve performance in dense retrieval. \citet{weller2025seq} further highlight that while bidirectional encoders excel at classification and retrieval tasks, the lack of large-scale encoder-only models in practice suggests that embedding models based on decoder-only LLMs will likely surpass all other options. Nevertheless, decoder-only LLM-based embedding methods still suffer from the inherent architectural drawbacks: causal attention prevents each token from interacting with subsequent tokens, hindering the model's ability to produce contextualized representations. To address this issue, BeLLM~\cite{li-li-2024-bellm} transforms LLM's attention from unidirectional to bidirectional by removing the causal mask at specific layers. GRITLM~\citep{grit} and LLM2Vec~\citep{llm2vec} further enable fully bidirectional attention in LLMs. Building upon this attention modification, NV-Embed~\citep{nvembed} introduces a latent attention layer over LLM's output hidden states to generate higher-quality representations. Moreover, MGH~\citep{mgh} proposes a novel pooling method that dynamically aggregates the output sequence to acquire a more accurate text embedding. 

Notably, these bidirectional LLM-based methods involve modifications to the model architecture, limiting their compatibility with diverse LLM backbones. In contrast, some studies preserve the original causal attention while attempting to alleviate its limitations. ~\citet{e5mistral} employ proprietary LLMs to construct diverse synthetic data and fint-tune open-source decoder-only LLMs through standard contrastive learning, achieving competitive performance. ECHO~\citep{echo} repeats the input twice in the autoregressive modeling paradigm, allowing the text embedding extracted from the repeated tokens to capture contextualized content. Similarly, TP~\cite{fu2025token} prepends the previous layer’s last-token to the next layer’s input sequence, enabling tokens to attend to the full sentence information. PromptEOL~\citep{prompteol} and bge-en-icl~\citep{bgeicl} enhance text embeddings by leveraging the in-context learning (ICL) capabilities of LLMs, augmenting the original input with task-specific examples to provide contextual information. Furthermore, Anchor~\cite{anchor} introduces a bidirectional reconstruction training stage before contrastive learning to enrich the semantics of the final embedding.

\begin{figure*}[htbp]
\centering
\includegraphics[width=\textwidth]{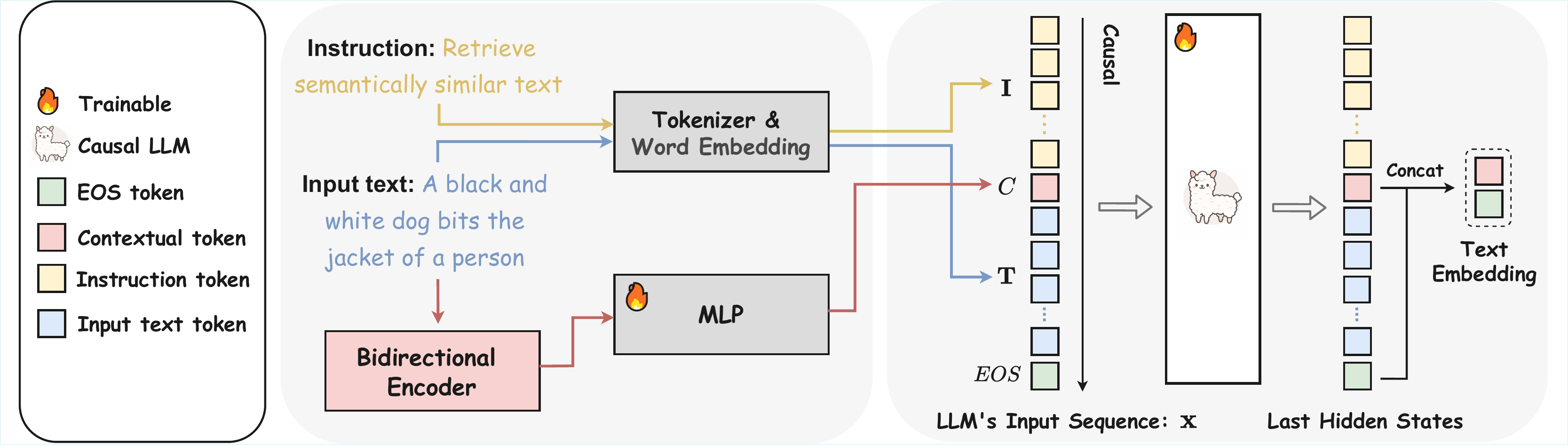}
\caption{Overview of our Causal2Vec method.}
\label{fig:model}
\end{figure*}
\begin{figure}[t]
    \centering
    \includegraphics[scale=0.33]{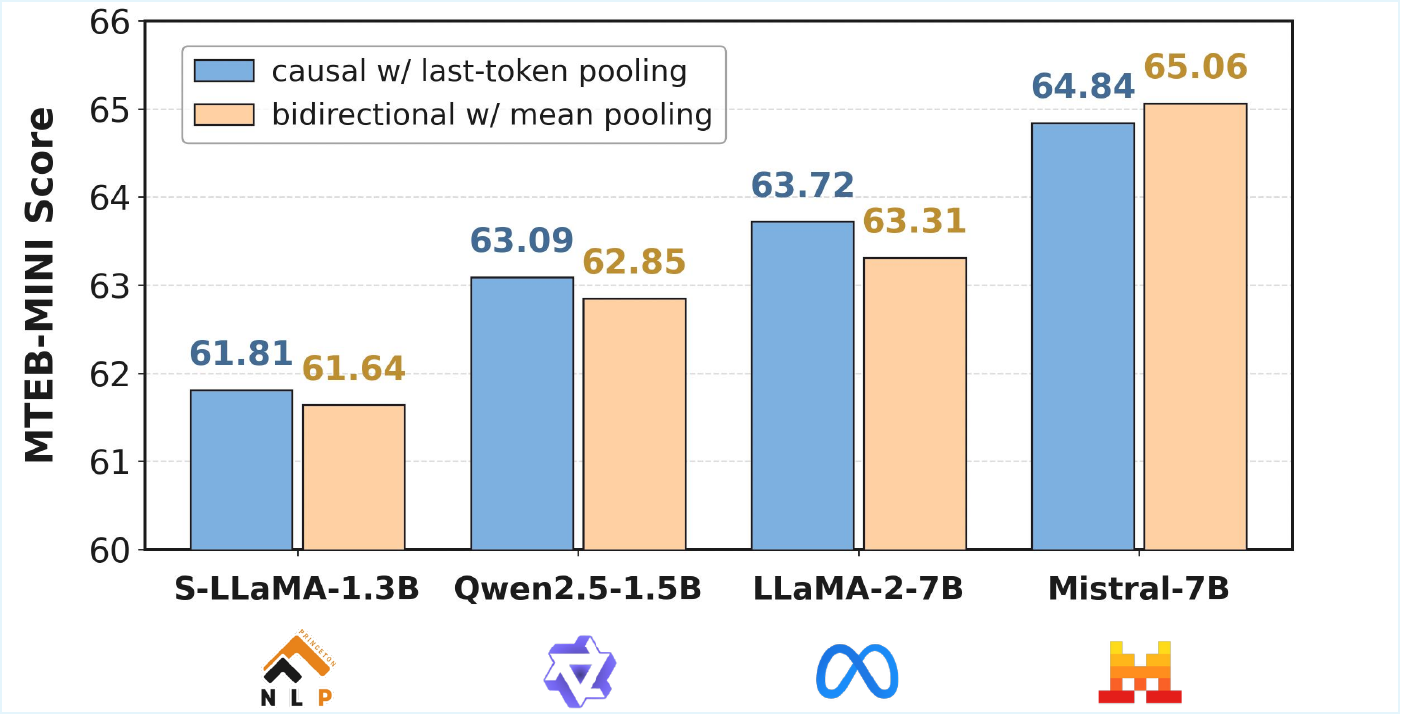}
    \caption{Average MTEB-MINI score (30 datasets) of causal vs. bidirectional attention across different LLM backbones. All results are obtained after contrastive learning on publicly available retrieval datasets.}
    \label{fig:causal_or_bi}
\end{figure}

\section{Method}
Figure~\ref{fig:model} illustrates an overall pipeline of the proposed Causal2Vec. Given an input text, we first use a lightweight bidirectional encoder to generate the Contextual token, which is then added to LLM's input sequence for causal attention computation.  The final text embedding is derived by concatenating the last hidden states of the Contextual and EOS tokens. We elaborate on the Contextual token and representation method in the following sections.
\label{main:method}
\subsection{Preserve the Causal Attention}
The remarkable capacity of LLMs for human language understanding and generation is acquired through training on massive amounts of text data~\citep{ouyang2022training,wei2022chain}, showcasing their effectiveness in encoding semantic information. However, the inherent causal attention in LLMs prevents earlier tokens from accessing information about future tokens, thus hindering the model’s representational capability. 

To mitigate this limitation, many LLM-based embedding models switch from unidirectional attention to bidirectional by removing the causal attention mask. While bidirectional attention facilitates effective information flow across the entire sentence, unlike models trained from scratch under bidirectional attention, this modification introduces an attention mismatch between pre-training and fine-tuning, potentially compromising the LLMs’ ability to extract semantic information acquired during pre-training. As shown in Figure~\ref{fig:causal_or_bi}, altering the attention mechanism degrades embedding performance across most LLMs—except for Mistral-7B~\citep{mistral}, likely due to its non-standard pre-training methodology~\citep{echo,llm2vec}. Consequently, most existing bidirectional methods~\citep{grit,nvembed,mgh} rely on Mistral-7B as the foundation model, limiting their generalizability to diverse LLM architectures. Furthermore, ~\citet{bgeicl} show that converting Mistral-7B to bidirectional harms its in-context learning capabilities for text embedding tasks. 

These findings suggest that many LLMs do not benefit from attention modification, making it neither a robust nor a general solution for transforming decoder-only LLMs into text embedding models. Consequently, we preserve the original causal attention and instead explore alternative strategies to overcome its representational bottleneck. Please refer to Appendix~\ref{appendix:different_attention} for more comparisons.

\subsection{The Contextual Token}
To fully unlock the potential of decoder-only LLMs in generating text embeddings, it is essential to address the limitations of causal attention while preserving LLMs' ability to extract well-learned semantic information. To this end, we first introduce a lightweight BERT-style model that encodes the input text into a $k$-dimensional dense vector representation $h\in \mathbb{R}^{1\times k}$, termed the "\emph{Contextual Token}". Specifically, this token is generated by applying mean pooling over the last hidden state of the additional bidirectional model, capturing contextualized information about the entire input. Furthermore, to bridge the gap between the BERT-style model and LLMs, we employ a simple MLP layer to match the dimensionality of the Contextual token with LLM’s word embedding space, and then encourage the LLMs to understand the sentence information encoded in this token through contrastive learning. Motivated by~\citet{llava2}, the MLP layer consists of two linear transformations with a GELU activation $\sigma$, which can be formulated as:
\vspace{-0.04cm}
\begin{equation}\label{equ1}\normalsize 
C=\sigma(h\boldsymbol{\mathrm{W}}_1^\top)\boldsymbol{\mathrm{W}}_2^\top ,
\end{equation}
where $\boldsymbol{\mathrm{W}}_1\in \mathbb{R}^{d \times k}$ and $\boldsymbol{\mathrm{W}}_2\in \mathbb{R}^{d \times d}$ are trainable projection matrices, and $C \in \mathbb{R}^{1\times d}$ denotes the language embedding of the Contextual token, which shares the same dimensionality $d$ as the word embedding space of the LLMs. Moreover, to leverage LLMs' instruction-follow capabilities for producing general-purpose text embeddings, we use task-specific instructions for both training and evaluation, following ~\citet{e5mistral,llm2vec,echo}. Consequently, by prepending the instruction and Contextual tokens and appending the EOS token to the input sequence, the resulting sequence fed into the LLMs can be constructed as:
\begin{equation}\label{equ2}
\boldsymbol{\mathrm{x}} = [\boldsymbol{\mathrm{I}}; C; \boldsymbol{\mathrm{T}}; EOS] \in \mathbb{R}^{(l+n+2)\times d},
\end{equation}
where $[\cdot; \cdot]$ denotes the vertical concatenation operation, $\boldsymbol{\mathrm{I}} \in \mathbb{R}^{l\times d}$ and $\boldsymbol{\mathrm{T}} \in \mathbb{R}^{n\times d}$ represent the word embeddings of the task-specific instruction and input text, respectively. 
In this way, each token following the Contextual token $C$ can capture contextualized information even without attending to future tokens. More importantly, the use of the Contextual token requires no modifications to the model architecture, which not only preserves LLMs' ability to extract semantic information learned during pre-training, but also enables seamless integration across different LLMs.

\subsection{Representation Method}
As the most widely adopted representation method for unidirectional models~\citep{e5mistral,bgeicl}, last-token pooling typically utilizes the final hidden state of the EOS token as text embeddings, since only the last token captures information from the entire input. However, recent studies~\citep{echo,nvembed} indicate that the EOS token embedding depends heavily on tokens near the end of the sequence, leading to potential semantic bias in long-text scenarios.

To address this issue, we introduce a simple yet effective representation method tailored for our embedding framework. Specifically, we concatenate the hidden states of the Contextual and EOS tokens from the LLM's last layer to generate text embeddings. Unlike last-token pooling, the Contextual token is not affected by tokens near the end of the sequence, thus effectively mitigating recency bias. In addition, since the Contextual token has already captured the semantic content of the input text, the concatenation of two context-aware tokens yields a vector representation with richer contextualized information. Furthermore, this approach enables explicit supervision of the Contextual token during training, thereby helping LLMs better understand the semantic information encoded in this added token. 
The proposed representation method is optimized through supervised contrastive learning with the standard InfoNCE loss~\citep{loss}, which can be formulated as:
\begin{equation}\label{equ3}\small
\mathcal{L} = -\log \frac{\exp(f(q, p^+)/ \tau)}{\exp(f(q, p^+)/ \tau) + \sum_{j=1}^{N} \exp(f(q, p_j^-)/ \tau)},\\
\end{equation}
where $f(q, p^+)$ represents the scoring function that computes cosine similarity between the query-passage pair embeddings from retrieval datasets, $p^-$ denotes both in-batch and hard negative examples, and $\tau$ is a temperature hyperparameter fixed at 0.05 in all experiments.

\section{Experiments}
\label{sec:experiments}
\subsection{Datasets}
\paragraph{Training Datasets.} For training, we follow the mainstream practices~\citep{llm2vec,anchor,mgh,bgeicl}, utilizing the same collection of publicly available retrieval datasets curated by~\citet{echo}, which consists of approximately 1.5 million samples. More details about the composition of the training datasets can be found in Appendix~\ref{appendix:training_data}.

Notably, many industry-developed embedding models, such as Qwen3 Embedding~\cite{zhang2025qwen3}, achieve strong performance by leveraging large amounts of non-public or proprietary synthetic data across various embedding tasks. To ensure fairness and consistency in academic comparisons, we evaluate only models trained on public retrieval datasets, enabling the verification of models' generalization capability to unseen non-retrieval tasks, which serves as a critical criterion for defining a general-purpose embedding model.

\begin{table*}[t]
    \centering
    \small
    \resizebox{\textwidth}{!}{
    \tabcolsep=3pt
    \begin{tabular}{lcccccccc}
    \toprule
    \textbf{Task (\# of datasets})\phantom{...} & \textbf{Retr. (15)} & \textbf{Rerank. (4)} & \textbf{Clust. (11)} & \textbf{PairClass. (3)} & \textbf{Class. (12)} & \textbf{STS (10)} & \textbf{Summ. (1)} & \textbf{Avg (56)} \\
    \textbf{Metric} & \multicolumn{1}{c}{nDCG@10}     & \multicolumn{1}{c}{MAP}     & \multicolumn{1}{c}{V-Meas.}   & \multicolumn{1}{c}{AP}      & \multicolumn{1}{c}{Acc.}    & \multicolumn{1}{c}{Spear.}  & \multicolumn{1}{c}{Spear.}    & \multicolumn{1}{c}{} \\ 
    \midrule
    \multicolumn{9}{c}
    {\textbf{\texttt{S-LLaMA-1.3B}}~~\raisebox{-0.2\height}{\includegraphics[height=1em]{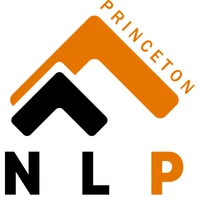}}}\\\midrule
    LLM2Vec~\citep{llm2vec} &  51.44 & 55.38 & 43.57 & 86.20 & 72.21 & 83.58 & 30.01 & 61.85 \\
    ECHO~\citep{echo} &  - & - & - & - & - & - & - & 62.01 \\
    \rowcolor{blue!10}Causal2Vec &  52.69 & 56.54 & 44.35 & 86.18 & 72.94 & 83.76 & \underline{31.45} & 62.63\\
    \midrule
    \multicolumn{9}{c}
    {\textbf{\texttt{Qwen2.5-1.5B}}~~\raisebox{-0.2\height}{\includegraphics[height=1em]{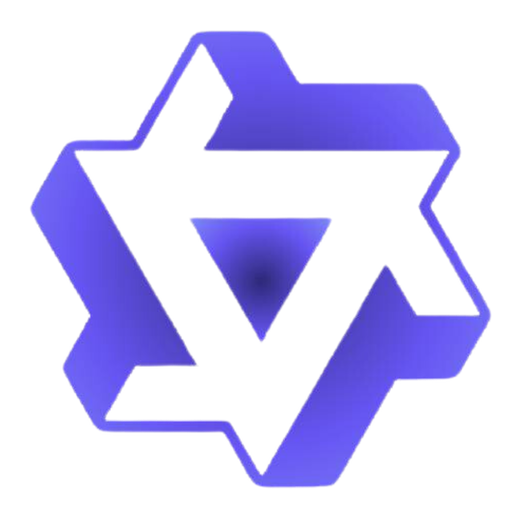}}}\\\midrule
    Anchor~\citep{anchor} & 53.62 & 57.63 & 43.19 & 85.77 & 74.51 & 82.74 & 31.61 &  62.86 \\
    \rowcolor{blue!10}Causal2Vec &  54.57 & 58.27 & 46.30 & 86.73 & 73.68 & 84.63 & 30.99 & 63.97 \\
    \midrule
    \multicolumn{9}{c}
    {\textbf{\texttt{LLaMA-2-7B}}~~\raisebox{-0.2\height}{\includegraphics[height=1em]{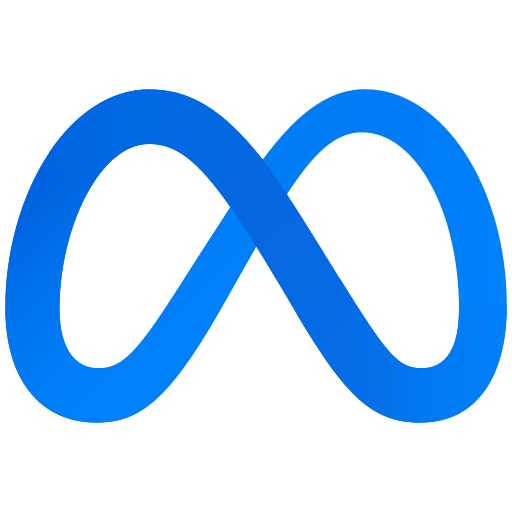}}}\\\midrule
    LLM2Vec~\citep{llm2vec} &  54.60 & 57.38 & 45.24 & 88.03 & 76.33 & 83.73 & 28.49 & 64.14 \\
    \rowcolor{blue!10}Causal2Vec &  55.28 & 58.18 & 47.23 & 87.85 & 75.95 & 84.90 & 31.09 & 64.94\\
    \midrule
    \multicolumn{9}{c}{\textbf{\texttt{Mistral-7B}}~~\raisebox{-0.1\height}{\includegraphics[height=1em]{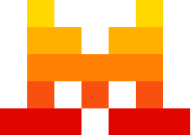}}}\\\midrule
    \multicolumn{1}{l}{E5$_\text{Mistral-7b}$~\citep{e5mistral}}  & 52.78 & 60.38 & 47.78 & 88.47 & 76.80 & 83.77 & \textbf{31.90} & 64.56 \\
    ECHO~\citep{echo} & 55.52 & 58.14 & 46.32 & 87.34 & 77.43 & 82.56 & 30.73 & 64.68\\
    GRITLM~\citep{grit} & 53.10 & \textbf{61.30} & \underline{48.90} & 86.90 & 77.00 & 82.80 & 29.40 & 64.70 \\
    LLM2Vec~\citep{llm2vec} & 55.99 & 58.42 & 45.54 & 87.99 & 76.63 & 84.09 & 29.96 & 64.80\\
    Anchor\citep{anchor} & 56.87 & \underline{60.56} & 45.73 & 87.99 & 75.95 & 83.52 & 30.28 & 64.99 \\
    NV-Embed\dag~\citep{nvembed} & - & - & - & - & - & - & - & 65.80\\
    MGH~\citep{mgh} & \underline{57.49} & 58.80 & 47.96 & 87.83 & \textbf{77.62} & 84.04 & 31.10 & 65.87\\
    bge-en-icl~\citep{bgeicl} & \textbf{60.08} & 56.67 & 46.55 & \underline{88.51} & 77.31 & 83.69 & 30.68 & 66.08\\
    \rowcolor{blue!10}Causal2Vec & 57.28 & 59.46 & 48.89 & 88.43 & 76.41 & \underline{85.38} & 30.57 & \underline{66.10}\\
    \rowcolor{blue!10}Causal2Vec w/ ICL & \cellcolor{blue!10}57.48 & 59.36 & \textbf{50.78} & \textbf{89.19} & \underline{77.53} & \textbf{85.66} & 30.82 & \textbf{66.85}\\
    \midrule
    \end{tabular}
    }
    \caption{Performance comparison on MTEB (56 datasets) 
    for models trained on publicly available retrieval datasets. \texttt{S-LLaMA-1.3B}, \texttt{Qwen2.5-1.5B}, \texttt{LLaMA-2-7B}, and \texttt{Mistral-7B} refer to embedding methods built upon these decoder-only LLMs. \dag~denotes results reported by \citet{mgh}. The best results are highlighted in \textbf{bold}, and the second-best are \underline{underlined}. See Appendix~\ref{appendix:full_mteb} for detailed results for each dataset.
    }
    \label{tab:mteb}
\end{table*}
\paragraph{Evaluation Benchmark.} For evaluation, we use the English subset of the Massive Text Embeddings Benchmark (MTEB)~\citep{mteb}, which comprises 56 datasets spanning 7 embedding task categories. 
Since MTEB is a large-scale benchmark containing over 35 million test samples, the full evaluation of a 7B parameter model is GPU resource intensive, requiring over 200 A100 80GB GPU hours. To speed up the evaluation, we curate a representative subset of MTEB, termed MTEB-MINI, for ablation
studies and analysis. Specifically, MTEB-MINI consists of 30 datasets covering all task categories in MTEB. Details of the MTEB-MINI composition can be found in Appendix~\ref{appendix:mteb-mini}.

\subsection{Experimental Details}
\label{sec:implementation}
For the base model, we integrate Causal2Vec with four decoder-only LLMs ranging from 1.3B to 7B parameters: Sheared-LLaMA-1.3B (S-LLaMA-1.3B), Qwen2.5-1.5B-Instruct (Qwen2.5-1.5B), Llama-2-7B-chat (LLaMA-2-7B), and Mistral-7B-Instruct-v0.2 (Mistral-7B). Regarding the off-the-shelf bidirectional encoder, we adopt E5-base-v2~\citep{e5}, a lightweight model with only 110M parameters. All LLMs are fine-tuned using LoRA~\cite{hu2022lora} on A100 80GB GPUs. In particular, LoRA is also applied to the bidirectional encoder when using the 1.3B and 1.5B LLMs (see Appendix~\ref{appendix:freeze_bert} for the reason). More experimental details are provided in Appendices~\ref{appnendix: details_training} and~\ref{appnendix: details_evaluation}.

\subsection{MTEB Results}
We evaluate the proposed Causal2Vec against competitive embedding models on the MTEB benchmark. As shown in Table~\ref{tab:mteb}, Causal2Vec demonstrates consistently strong performance across various LLM backbones, with our best model, Causal2Vec-Mistral-7B, achieving new state-of-the-art results compared to methods trained solely on publicly available retrieval datasets.

\textbf{Comparison to Bidirectional LLM-based Methods.}
GRITLM~\citep{grit} and LLM2Vec~\citep{llm2vec} remove the causal attention mask to enable bidirectional embedding generation. Building on this attention modification, NV-Embed~\citep{nvembed} introduces a latent attention layer to obtain pooled embeddings, while MGH~\citep{mgh} initializes from LLM2Vec and leverages LLMs' inherent aggregation patterns to derive stronger embeddings over mean pooling. In contrast to these bidirectional LLM-based methods, our Causal2Vec requires no modifications to the model architecture, yet enables each token in the sequence to access contextual information through the introduced Contextual token. More importantly, our approach surpasses all aforementioned bidirectional models on MTEB. Specifically, Causal2Vec consistently outperforms LLM2Vec across three LLM base models. Notably, Causal2Vec exceeds the state-of-the-art bidirectional methods MGH (66.10 vs. 65.87) under identical training data and evaluation pipelines. These results underscore that shifting from causal attention to bidirectional is not necessary for adopting LLMs in text embedding tasks—and may even compromise the model’s ability to extract the well-learned semantic information. We argue that the effectiveness of bidirectional attention in capturing contextual information relies on maintaining consistency in attention mechanisms throughout both pre-training and fine-tuning. 

\textbf{Comparison to Causal LLM-based Methods.}
Anchor~\citep{anchor} enhances the EOS representation through two additional reconstruction objectives, but requires extensive full-parameter tuning. In comparison, our Causal2Vec outperforms Anchor by $1.11$ points on both Qwen-2.5-1.5B and Mistral-7B, using only standard contrastive learning with LoRA tuning.

ECHO~\citep{echo} repeats the input, allowing each token from the second occurrence to access the full sequence. However, this strategy doubles the maximum sequence length, inevitably increasing computational cost. In contrast, our Causal2Vec enables contextualized information access through a single Contextual token, and consistently outperforms ECHO on S-LLaMA-1.3B and Mistral-7B by $0.62$ and $1.42$ points, respectively.

The state-of-the-art unidirectional method bge-en-icl~\citep{bgeicl} endows LLM-based embedding models with in-context learning (ICL)~\citep{icl} capabilities by incorporating multiple task-related examples into the input. Causal2Vec achieves embedding performance on par with bge-en-icl on MTEB (66.10 vs. 66.08). However, incorporating in-context examples substantially increases the computational burden of LLMs, especially given that the maximum sequence length of bge-en-icl can reach up to 2048 tokens, which is four times that of our method. More importantly, when equipped with the same ICL strategy, Causal2Vec (w/ ICL) achieves a new state-of-the-art score of \textbf{66.85}, outperforming bge-en-icl by 0.77 points under the compute-matched setting.

\begin{figure}[t]
    \centering
    \includegraphics[scale=0.43]{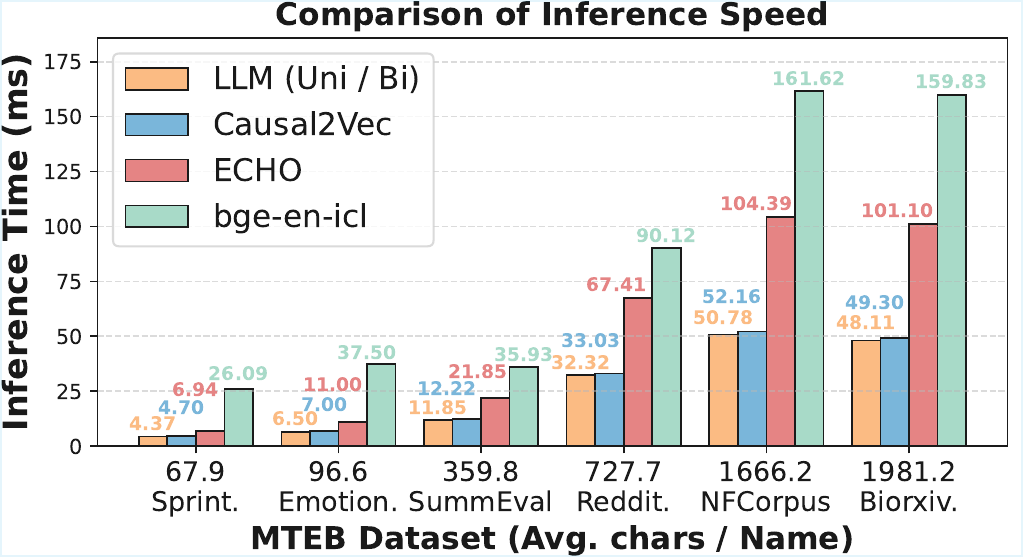}
    \caption{Average inference time per sample (in milliseconds) for various Mistral-7B-based methods on selected MTEB subsets, evaluated with batch size of 32 on a single NVIDIA A100 80GB GPU. \texttt{LLM (Uni/Bi)} denotes the standard Mistral-7B with causal or bidirectional attention. For the asymmetric dataset NFCorpus, we report the results per query-passage pair.}
    \label{fig:inference_comparison}
\end{figure}

\textbf{Efficiency.} In Figure~\ref{fig:inference_comparison}, we report the approximate average inference time for different methods. Specifically, ECHO and bge-en-icl inevitably increase inference time due to their extended sequence lengths. In contrast, since the additional bidirectional encoder contains only 110M parameters and a single Contextual token adds no computational burden to the LLM, our Causal2Vec achieves inference speed comparable to the standard Mistral-7B using causal or bidirectional attention. Notably, compared to the previous best-performing method, beg-en-icl, Causal2Vec reduces inference time by up to \textbf{82\%} (e.g., Sprint.: $4.37$ vs. $26.09$; Emotion.: $6.50$ vs. $37.50$). See Appendix~\ref{appnendix: length_comparison} for a comparison of the required sequence length.

\definecolor{myred}{RGB}{220,50,47} 
\definecolor{mygreen}{RGB}{133,153,0}
\newcommand{\cmark}{\textcolor{mygreen}{\ding{51}}}  
\newcommand{\xmark}{\textcolor{myred}{\ding{55}}} 

\begin{table}[t]
    \centering
    \scalebox{0.85}{
    \begin{tabular}{cccc}
    \toprule
    \textbf{CtxToken} & \textbf{Concat} & \textbf{S-LLaMA-1.3B} &  \textbf{Mistral-7B} \\
    \midrule
    \xmark & \xmark &  61.81  & 64.84\\
    \cmark & \xmark &  62.19  &  65.44 \\
    \rowcolor{blue!10}\cmark & \cmark &  \textbf{62.83} & \textbf{65.85} \\
    \midrule
    \end{tabular}}
    \caption{Performance comparison of different components on MTEB-MINI (30 datasets). \texttt{CtxToken} indicates adding the Contextual token to LLM's input, while \texttt{Concat} denotes the proposed representation method that concatenates the last hidden states of LLM's Contextual and EOS tokens as the text embedding.}
    \label{tab:component}
\end{table}

\begin{table}[t]
\centering
\scalebox{0.85}{
\setlength{\tabcolsep}{9pt} 
\begin{tabular}{l cc}
\toprule
\textbf{Method} & \textbf{Retr. (6)} & \textbf{MTEB-MINI (30)} \\
\midrule
Mistral-7B  &  49.44 & 64.84 \\
\quad w/ noise & 44.02 (\textbf{\textcolor[rgb]{0.86,0.20,0.18}{-5.42}}) & 58.69 (\textbf{\textcolor[rgb]{0.86,0.20,0.18}{-6.15}}) \\
\midrule
Causal2Vec & 50.81 & 65.85 \\
\quad w/ noise & 47.56 (\textcolor[rgb]{0.86,0.20,0.18}{-3.25}) & 60.95 (\textcolor[rgb]{0.86,0.20,0.18}{-4.90})\\
\bottomrule 
\end{tabular}
}
\caption{Performance comparison under input noise for retrieval tasks (6 datasets) and MTEB-MINI (30 datasets), between standard Mistral-7B with last-token pooling and Causal2Vec-Mistral-7B.}
\label{tab:noise_input}
\end{table}

\subsection{Ablation Studies}
\label{main:ablation}
\paragraph{Effectiveness of Each Component.}
To evaluate the effectiveness of the proposed Contextual token and representation method, we conduct ablation studies on MTEB-MINI using two base models of different scales: S-LLaMA-1.3B and Mistral-7B. As shown in Table~\ref{tab:component}, incorporating the Contextual token into the LLM's causal attention mechanism yields average score improvements of $0.38$ and $0.60$ for S-LLaMA-1.3B and Mistral-7B, respectively. These results not only confirm the effectiveness of the Contextual token but also highlight its scalability and applicability across different LLMs. We attribute these performance improvements to the rich contextualized content encoded in the Contextual token, which allows preceding tokens in the sequence to access accurate sentence information even without attending to future tokens. This mitigates the inherent architectural limitations in causal attention while preserving the original autoregressive paradigm.

By concatenating the last hidden states of the Contextual and EOS tokens as the final vector representation, we observe consistent performance improvements across all seven tasks, with average score gains of $1.02$ and $1.01$ on S-LLaMA-1.3B and Mistral-7B compared to standard LLMs, respectively. To further evaluate the robustness of our representation method in alleviating recency bias, we append task-irrelevant text to LLM's input sequence as noise. As shown in Table~\ref{tab:noise_input}, standard Mistral-7B with last-token pooling suffers a significant average performance drop of $6.15$ points, while Causal2Vec exhibits a smaller degradation of $4.90$. Notably, the gap is larger in retrieval tasks, where the drops are $5.42$ vs. $3.25$, revealing the high sensitivity of last-token pooling to noise in long-text scenarios. See Appendix~\ref{appendix:representation_type} for further discussion on various representation methods.

To verify whether the performance gains of our method arise merely from embedding fusion, we compare Causal2Vec with a baseline that directly concatenates the output representations from the LLM and the bidirectional encoder. As shown in Table~\ref{tab:compare_direct_cocnat}, this simple strategy leads to significant performance degradation, suggesting that the improvements achieved by Causal2Vec stem from addressing the inherent shortcomings of causal attention and last-token 
pooling.

\begin{table}[t]
\centering
\scalebox{0.8}{
\begin{tabular}{l cc}
\toprule
\textbf{Method} & \textbf{S-LLaMA-1.3B}  & \textbf{Mistral-7B} \\
\midrule
Embedding Concatenation & 61.92 & 65.02 \\
\rowcolor{blue!10}Causal2Vec & \textbf{62.83}  &\textbf{65.85}  \\
\bottomrule 
\end{tabular}
}
\caption{Performance comparison between Causal2Vec and direct embedding concatenation on MTEB-MINI (30 datasets).}
\label{tab:compare_direct_cocnat}
\end{table}

\begin{table}[t]
\centering
\scalebox{0.8}{
\setlength{\tabcolsep}{10pt} 
\begin{tabular}{l cc}
\toprule
\textbf{Method} & \textbf{S-LLaMA-1.3B} &\textbf{Mistral-7B}\\
\midrule
\rowcolor{blue!10}Causal2Vec &  \textbf{62.83} & \textbf{65.85}\\
\quad w/ 2 CtxTokens & 62.77 & 65.76\\
\quad w/ 4 CtxTokens & 62.51 & 65.72\\
\quad w/ 8 CtxTokens & 62.67 & 65.68\\
\bottomrule 
\end{tabular}
}
\caption{Performance comparison of Causal2Vec using different numbers of Contextual tokens (\texttt{CtxTokens}) on MTEB-MINI (30 datasets). Note: Causal2Vec uses a single Contextual token by default.}
\label{tab:number_of_contextual}
\end{table}

\paragraph{The Number of the Contextual Tokens.}
To examine the impact of using multiple Contextual tokens, we adopt cross-attention with a set of learnable queries to extract a fixed number of Contextual tokens from the bidirectional encoder, following~\citet{detr,blip2}. As presented in Table~\ref{tab:number_of_contextual}, increasing the number of the Contextual tokens leads to performance
degradation. We hypothesize that the additional tokens fail to provide more distinctive semantic information and instead introduce redundancy. These findings suggest that a single Contextual token is sufficient to supply the necessary contextual information for Causal2Vec, while maintaining model simplicity and efficiency.

\paragraph{Impact of Different Bidirectional Encoders.}
Finally, we explore the impact of using different bidirectional encoders in Causal2Vec, including E5-small-v2, E5-base-v2, E5-large-v2~\cite{e5}, and GTE-small~\cite{gte}, which achieve standalone MTEB scores of $59.93$, $60.97$, $61.44$, and $61.36$, respectively. As shown in Table~\ref{tab:bert_version}, incorporating the Contextual token consistently outperforms the baseline LLM with last-token pooling (\texttt{w/o Bi-Encoder}), confirming the robustness of Causal2Vec across various bidirectional encoders. Notably, replacing the default E5-base-v2 with stronger encoders such as E5-large-v2 and GTE-small does not yield further meaningful performance gains. These findings suggest that the performance ceiling of our architecture is primarily determined by the LLM itself rather than the additional bidirectional encoder.

In addition, we compare Causal2Vec with TP~\cite{fu2025token}, which prepends each layer’s output EOS token to the input sequence of next layer. Our method substantially outperforms TP, demonstrating that compared to repeatedly using EOS token within the LLM, our Contextual token generated by the additional encoder is a more effective solution for providing contextualized information.
\begin{table}[t]
\centering
\scalebox{0.72}{
\begin{tabular}{l ccc}
\toprule
\textbf{Method} & \textbf{Params.} & \textbf{ S-LLaMA-1.3B } & \textbf{Mistral-7B} \\
\midrule
\rowcolor{blue!10} Causal2Vec  & 110M & \textbf{62.83} & \textbf{65.85} \\
\quad w/ GTE-small & 33M & 62.71 & 65.45 \\
\quad w/ E5-small-v2  & 33M & 62.70 &65.41\\
\quad w/ E5-large-v2 & 335M & 62.31 & 65.71\\
\quad w/o Bi-Encoder & N/A & 61.81 & 64.84 \\
TP~\cite{fu2025token} & N/A & 61.66 & 64.93 \\ 
\bottomrule 
\end{tabular}
}
\caption{Performance comparison of Causal2Vec using different bidirectional encoders (Bi-Encoders) on MTEB-MINI (30 datasets). Note: E5-base-v2~\cite{e5} is used as the default Bi-Encoder. "Params." indicates the parameter count of the Bi-Encoder.}
\label{tab:bert_version}
\end{table}

\section{Conclusion}
This paper presented Causal2Vec, a simple yet powerful text embedding model built upon decoder-only LLMs. It requires no architectural modifications or additional input text, achieving consistently strong performance in text embedding tasks. By introducing the Contextual token, we enable each token in the sequence to capture contextual information within the inherent autoregressive modeling paradigm. To address the limitations of last-token pooling, commonly used in unidirectional models, we proposed a specialized representation method that concatenates the last hidden states of the Contextual and EOS tokens as the final text embedding. Experimental results demonstrated that Causal2Vec achieves new state-of-the-art performance on MTEB across different LLM backbones. Extensive ablations and analyses further confirmed the effectiveness of the proposed mechanism. 

\section*{Limitations}
Despite the effectiveness of Causal2Vec, several limitations should be acknowledged: (1) Our findings suggest that a single Contextual token is sufficient to provide the missing contextual information for decoder-only LLMs. Future work could explore generating additional Contextual tokens using different bidirectional encoders or utilizing multiple task-related examples. (2) Our experiments are limited to four popular LLMs with fewer than 7B parameters, while further validation on more diverse and larger-scale LLMs could better demonstrate the scalability and robustness of our mechanism. (3) Existing state-of-the-art embedding methods~\cite{llm2vec,echo,nvembed,mgh} primarily focus on evaluations using English benchmarks. We plan to further explore the performance of Causal2Vec on multilingual text embedding tasks. (4) The representation method used in Causal2Vec doubles the output embedding dimension, which may increase storage overhead in certain scenarios.

\bibliography{main}

@inproceedings{ouyang2022training,
  title={Training language models to follow instructions with human feedback},
  author={Ouyang, Long and Wu, Jeffrey and Jiang, Xu and Almeida, Diogo and Wainwright, Carroll L. and Mishkin, Pamela and Zhang, Chong and Agarwal, Sandhini and Slama, Katarina and Ray, Alex and others},
  booktitle={Advances in neural information processing systems},
  volume={35},
  pages={27730--27744},
  year={2022}
}

@article{wei2022chain,
  title={Chain-of-Thought Prompting Elicits Reasoning in Large Language Models},
  author={Wei, Jason and Wang, Xuezhi and Schuurmans, Dale and Bosma, Maarten and Ichter, Brian and Xia, Fei and Chi, Ed H and Le, Quoc V and Zhou, Denny},
  journal={Advances in Neural Information Processing Systems},
  volume={35},
  pages={24824--24837},
  year={2022}
}

@inproceedings{text2,
  title={Supervised learning of universal sentence representations from natural language inference data},
  author={Conneau, Alexis and Kiela, Douwe and Schwenk, Holger and Barrault, Lo{\"\i}c and Bordes, Antoine},
  booktitle={Proceedings of the 2017 Conference on Empirical Methods in Natural Language Processing},
  pages={670--680},
  year={2017}
}

@inproceedings{mteb,
  title={MTEB: Massive text embedding benchmark},
  author={Muennighoff, Niklas and Tazi, Nouamane and Magne, Lo{\"\i}c and Reimers, Nils},
  booktitle={Proceedings of the 17th Conference of the European Chapter of the Association for Computational Linguistics},
  pages={2014--2037},
  year={2023}
}

@inproceedings{bert,
  title={Bert: Pre-training of deep bidirectional transformers for language understanding},
  author={Devlin, Jacob and Chang, Ming-Wei and Lee, Kenton and Toutanova, Kristina},
  booktitle={Proceedings of the 2019 conference of the North American chapter of the association for computational linguistics: human language technologies, volume 1 (long and short papers)},
  pages={4171--4186},
  year={2019}
}

@article{t5,
  title={Exploring the limits of transfer learning with a unified text-to-text transformer},
  author={Raffel, Colin and Shazeer, Noam and Roberts, Adam and Lee, Katherine and Narang, Sharan and Matena, Michael and Zhou, Yanqi and Li, Wei and Liu, Peter J},
  journal={Journal of Machine Learning Research},
  volume={21},
  number={140},
  pages={1--67},
  year={2020}
}

@inproceedings{karpukhin2020dense,
  title={Dense Passage Retrieval for Open-Domain Question Answering.},
  author={Karpukhin, Vladimir and Oguz, Barlas and Min, Sewon and Lewis, Patrick and Wu, Ledell and Edunov, Sergey and Chen, Danqi and Yih, Wen-tau},
  booktitle={Proceedings of the 2020 Conference on Empirical Methods in Natural Language Processing (EMNLP)},
  pages={6769--6781},
  year={2020}
}

@inproceedings{xiong2021approximate,
 title={Approximate Nearest Neighbor Negative Contrastive Learning for Dense Text Retrieval},
 author={Lee Xiong and Chenyan Xiong and Ye Li and Kwok-Fung Tang and Jialin Liu and Paul N. Bennett and Junaid Ahmed and Arnold Overwijk},
 booktitle={International Conference on Learning Representations},
 year={2021}
}

@inproceedings{liu2024chatqa,
 title={Chat{QA}: Surpassing {GPT}-4 on Conversational {QA} and {RAG}},
 author={Zihan Liu and Wei Ping and Rajarshi Roy and Peng Xu and Chankyu Lee and Mohammad Shoeybi and Bryan Catanzaro},
 booktitle = {Advances in Neural Information Processing Systems},
 year={2024}
}

@inproceedings{llm2vec,
 title={{LLM}2Vec: Large Language Models Are Secretly Powerful Text Encoders},
 author={Parishad BehnamGhader and Vaibhav Adlakha and Marius Mosbach and Dzmitry Bahdanau and Nicolas Chapados and Siva Reddy},
 booktitle={First Conference on Language Modeling},
 year={2024}
}

@inproceedings{nvembed,
 title={{NV}-Embed: Improved Techniques for Training {LLM}s as Generalist Embedding Models},
 author={Chankyu Lee and Rajarshi Roy and Mengyao Xu and Jonathan Raiman and Mohammad Shoeybi and Bryan Catanzaro and Wei Ping},
 booktitle={The Thirteenth International Conference on Learning Representations},
 year={2025}
}

@inproceedings{grit,
  title={Generative Representational Instruction Tuning},
  author={Muennighoff, Niklas and Hongjin, SU and Wang, Liang and Yang, Nan and Wei, Furu and Yu, Tao and Singh, Amanpreet and Kiela, Douwe},
  booktitle={ICLR 2024 Workshop: How Far Are We From AGI},
  year={2024}
}

@inproceedings{echo,
title={Repetition Improves Language Model Embeddings},
author={Jacob Mitchell Springer and Suhas Kotha and Daniel Fried and Graham Neubig and Aditi Raghunathan},
booktitle={The Thirteenth International Conference on Learning Representations},
year={2025}
}

@inproceedings{bgeicl,
title={Making Text Embedders Few-Shot Learners},
author={Chaofan Li and Minghao Qin and Shitao Xiao and Jianlyu Chen and Kun Luo and Defu Lian and Yingxia Shao and Zheng Liu},
booktitle={The Thirteenth International Conference on Learning Representations},
year={2025}
}

@inproceedings{icl,
  title={Language models are few-shot learners},
  author={Brown , Tom B. and Mann, Benjamin and Ryder, Nick and Subbiah, Melanie and Kaplan, Jared and Dhariwal, Prafulla and Neelakantan, Arvind and Shyam, Pranav and Sastry, Girish and Askell, Amanda and others},
  booktitle={Advances in Neural Information Processing Systems},
  volume={33},
  pages={1877--1901},
  year={2020}
}

@article{roberta,
  title={Roberta: A robustly optimized bert pretraining approach},
  author={Liu, Yinhan and Ott, Myle and Goyal, Naman and Du, Jingfei and Joshi, Mandar and Chen, Danqi and Levy, Omer and Lewis, Mike and Zettlemoyer, Luke and Stoyanov, Veselin},
  journal={arXiv preprint arXiv:1907.11692},
  year={2019}
}

@inproceedings{simcse,
    title = "{S}im{CSE}: Simple Contrastive Learning of Sentence Embeddings",
    author = "Gao, Tianyu  and
      Yao, Xingcheng  and
      Chen, Danqi",
    booktitle = "Proceedings of the 2021 Conference on Empirical Methods in Natural Language Processing",
    year = "2021",
    pages = "6894--6910"
}

@inproceedings{st5,
    title = "Sentence-T5: Scalable Sentence Encoders from Pre-trained Text-to-Text Models",
    author = "Ni, Jianmo  and
      Hernández Ábrego, Gustavo  and
      Constant, Noah  and
      Ma, Ji  and
      B. Hall, Keith  and
      Cer, Daniel  and
      Yang, Yinfei",
    booktitle = "Findings of the Association for Computational Linguistics: ACL 2022",
    year = "2022",
    pages = "1864--1874"
}

@article{e5,
  title={Text Embeddings by Weakly-Supervised Contrastive Pre-training},
  author={Wang, Liang and Yang, Nan and Huang, Xiaolong and Jiao, Binxing and Yang, Linjun and Jiang, Daxin and Majumder, Rangan and Wei, Furu},
  journal={arXiv preprint arXiv:2212.03533},
  year={2022}
}

@article{gte,
  title={Towards general text embeddings with multi-stage contrastive learning},
  author={Li, Zehan and Zhang, Xin and Zhang, Yanzhao and Long, Dingkun and Xie, Pengjun and Zhang, Meishan},
  journal={arXiv preprint arXiv:2308.03281},
  year={2023}
}

@inproceedings{instrut1,
    title = "Task-aware Retrieval with Instructions",
    author = "Asai, Akari  and
      Schick, Timo  and
      Lewis, Patrick  and
      Chen, Xilun  and
      Izacard, Gautier  and
      Riedel, Sebastian  and
      Hajishirzi, Hannaneh  and
      Yih, Wen-tau",
    booktitle = "Findings of the Association for Computational Linguistics: ACL 2023",
    year = "2023",
    pages = "3650--3675"
}

@inproceedings{instrut2,
    title = "One Embedder, Any Task: Instruction-Finetuned Text Embeddings",
    author = "Su, Hongjin  and
      Shi, Weijia  and
      Kasai, Jungo  and
      Wang, Yizhong  and
      Hu, Yushi  and
      Ostendorf, Mari  and
      Yih, Wen-tau  and
      Smith, Noah A.  and
      Zettlemoyer, Luke  and
      Yu, Tao",
    booktitle = "Findings of the Association for Computational Linguistics: ACL 2023",
    year = "2023",
    pages = "1102--1121"
}

@article{instrut3,
  title={Multilingual e5 text embeddings: A technical report},
  author={Wang, Liang and Yang, Nan and Huang, Xiaolong and Yang, Linjun and Majumder, Rangan and Wei, Furu},
  journal={arXiv preprint arXiv:2402.05672},
  year={2024}
}

@inproceedings{luo-etal-2024-large,
    title = "Large Language Models as Foundations for Next-Gen Dense Retrieval: A Comprehensive Empirical Assessment",
    author = "Luo, Kun  and
      Qin, Minghao  and
      Liu, Zheng  and
      Xiao, Shitao  and
      Zhao, Jun  and
      Liu, Kang",
    booktitle = "Proceedings of the 2024 Conference on Empirical Methods in Natural Language Processing",
    year = "2024",
    pages = "1354--1365"
}

@inproceedings{prompteol,
    title = "Scaling Sentence Embeddings with Large Language Models",
    author = "Jiang, Ting  and
      Huang, Shaohan  and
      Luan, Zhongzhi  and
      Wang, Deqing  and
      Zhuang, Fuzhen",
    booktitle = "Findings of the Association for Computational Linguistics: EMNLP 2024",
    year = "2024",
    pages = "3182--3196"
}

@inproceedings{e5mistral,
    title = "Improving Text Embeddings with Large Language Models",
    author = "Wang, Liang  and
      Yang, Nan  and
      Huang, Xiaolong  and
      Yang, Linjun  and
      Majumder, Rangan  and
      Wei, Furu",
    booktitle = "Proceedings of the 62nd Annual Meeting of the Association for Computational Linguistics",
    year = "2024",
    pages = "11897--11916"
}

@inproceedings{llava2,
  title={Improved baselines with visual instruction tuning},
  author={Liu, Haotian and Li, Chunyuan and Li, Yuheng and Lee, Yong Jae},
  booktitle={Proceedings of the IEEE/CVF Conference on Computer Vision and Pattern Recognition},
  pages={26296--26306},
  year={2024}
}

@article{zhang2023language,
  title={Language models are universal embedders},
  author={Zhang, Xin and Li, Zehan and Zhang, Yanzhao and Long, Dingkun and Xie, Pengjun and Zhang, Meishan and Zhang, Min},
  journal={arXiv preprint arXiv:2310.08232},
  year={2023}
}

@article{loss,
  title={Unsupervised dense information retrieval with contrastive learning},
  author={Izacard, Gautier and Caron, Mathilde and Hosseini, Lucas and Riedel, Sebastian and Bojanowski, Piotr and Joulin, Armand and Grave, Edouard},
  journal={arXiv preprint arXiv:2112.09118},
  year={2021}
}

@inproceedings{sheared,
title={Sheared {LL}a{MA}: Accelerating Language Model Pre-training via Structured Pruning},
author={Mengzhou Xia and Tianyu Gao and Zhiyuan Zeng and Danqi Chen},
booktitle={The Twelfth International Conference on Learning Representations},
year={2024},
url={https://openreview.net/forum?id=09iOdaeOzp}
}

@article{llama2,
  title={Llama 2: Open foundation and fine-tuned chat models},
  author={Touvron, Hugo and Martin, Louis and Stone, Kevin and Albert, Peter and Almahairi, Amjad and Babaei, Yasmine and Bashlykov, Nikolay and Batra, Soumya and Bhargava, Prajjwal and Bhosale, Shruti and others},
  journal={arXiv preprint arXiv:2307.09288},
  year={2023}
}

@article{mistral,
  title={Mistral 7B},
  author={Jiang, Albert Q and Sablayrolles, Alexandre and Mensch, Arthur and Bamford, Chris and Chaplot, Devendra Singh and Casas, Diego de las and Bressand, Florian and Lengyel, Gianna and Lample, Guillaume and Saulnier, Lucile and others},
  journal={arXiv preprint arXiv:2310.06825},
  year={2023}
}

@inproceedings{hu2022lora,
title={Lo{RA}: Low-Rank Adaptation of Large Language Models},
author={Edward J Hu and yelong shen and Phillip Wallis and Zeyuan Allen-Zhu and Yuanzhi Li and Shean Wang and Lu Wang and Weizhu Chen},
booktitle={International Conference on Learning Representations},
year={2022},
url={https://openreview.net/forum?id=nZeVKeeFYf9}
}

@inproceedings{flashattention,
title={FlashAttention-2: Faster Attention with Better Parallelism and Work Partitioning},
author={Tri Dao},
booktitle={The Twelfth International Conference on Learning Representations},
year={2024},
url={https://openreview.net/forum?id=mZn2Xyh9Ec}
}

@article{llava,
  title={Visual instruction tuning},
  author={Liu, Haotian and Li, Chunyuan and Wu, Qingyang and Lee, Yong Jae},
  journal={Advances in Neural Information Processing Systems},
  volume={36},
  pages={34892--34916},
  year={2023}
}

@inproceedings{eli5,
    title = "{ELI}5: Long Form Question Answering",
    author = "Fan, Angela  and
      Jernite, Yacine  and
      Perez, Ethan  and
      Grangier, David  and
      Weston, Jason  and
      Auli, Michael",
    booktitle = "Proceedings of the 57th Annual Meeting of the Association for Computational Linguistics",
    year = "2019",
    pages = "3558--3567"
}

@inproceedings{hotpotqa,
    title = "{H}otpot{QA}: A Dataset for Diverse, Explainable Multi-hop Question Answering",
    author = "Yang, Zhilin  and
      Qi, Peng  and
      Zhang, Saizheng  and
      Bengio, Yoshua  and
      W. Cohen, William  and
      Salakhutdinov, Ruslan  and
      Manning, Christopher D.",
    booktitle = "Proceedings of the 2018 Conference on Empirical Methods in Natural Language Processing",
    year = "2018",
    pages = "2369--2380"
}

@inproceedings{fever,
    title = "{FEVER}: a Large-scale Dataset for Fact Extraction and {VER}ification",
    author = "Thorne, James  and
      Vlachos, Andreas  and
      Christodoulopoulos, Christos  and
      Mittal, Arpit",
    booktitle = "Proceedings of the 2018 Conference of the North {A}merican Chapter of the Association for Computational Linguistics: Human Language Technologies, Volume 1 (Long Papers)",
    year = "2018",
    pages = "809--819"
}

@article{miracl,
    title = "{MIRACL}: A Multilingual Retrieval Dataset Covering 18 Diverse Languages",
    author = "Zhang, Xinyu  and
      Thakur, Nandan  and
      Ogundepo, Odunayo  and
      Kamalloo, Ehsan  and
      Alfonso-Hermelo, David  and
      Li, Xiaoguang  and
      Liu, Qun  and
      Rezagholizadeh, Mehdi  and
      Lin, Jimmy",
    journal = "Transactions of the Association for Computational Linguistics",
    volume = "11",
    year = "2023",
    pages = "1114--1131"
}

@misc{marco,
title={{MS} {MARCO}: A Human-Generated {MA}chine Reading {CO}mprehension Dataset},
author={Tri Nguyen and Mir Rosenberg and Xia Song and Jianfeng Gao and Saurabh Tiwary and Rangan Majumder and Li Deng},
year={2017},
url={https://openreview.net/forum?id=Hk1iOLcle}
}

@inproceedings{MrTyDi,
    title = "Mr. {T}y{D}i: A Multi-lingual Benchmark for Dense Retrieval",
    author = "Zhang, Xinyu  and
      Ma, Xueguang  and
      Shi, Peng  and
      Lin, Jimmy",
    booktitle = "Proceedings of the 1st Workshop on Multilingual Representation Learning",
    year = "2021",
    pages = "127--137"
}

@inproceedings{dureader,
    title = "{D}u{R}eader: a {C}hinese Machine Reading Comprehension Dataset from Real-world Applications",
    author = "He, Wei  and
      Liu, Kai  and
      Liu, Jing  and
      Lyu, Yajuan  and
      Zhao, Shiqi  and
      Xiao, Xinyan  and
      Liu, Yuan  and
      Wang, Yizhong  and
      Wu, Hua  and
      She, Qiaoqiao  and
      Liu, Xuan  and
      Wu, Tian  and
      Wang, Haifeng",
    booktitle = "Proceedings of the Workshop on Machine Reading for Question Answering",
    year = "2018",
    publisher = "Association for Computational Linguistics",
    url = "https://aclanthology.org/W18-2605/",
    pages = "37--46"
}

@inproceedings{t2ranking,
  title={T2ranking: A large-scale chinese benchmark for passage ranking},
  author={Xie, Xiaohui and Dong, Qian and Wang, Bingning and Lv, Feiyang and Yao, Ting and Gan, Weinan and Wu, Zhijing and Li, Xiangsheng and Li, Haitao and Liu, Yiqun and Ma, Jin},
  booktitle={Proceedings of the 46th International ACM SIGIR Conference on Research and Development in Information Retrieval},
  pages={2681--2690},
  year={2023}
}

@article{quora,
    author = {DataCanary and hilfialkaff and Jiang, Lili and Risdal, Meg and Dandekar, Nikhil and tomtung},
    title = {Quora Question Pairs},
    year = {2017},
    url = {https://kaggle.com/competitions/quora-question-pairs}
}

@inproceedings{squad,
    title = "{SQ}u{AD}: 100,000+ Questions for Machine Comprehension of Text",
    author = "Rajpurkar, Pranav  and
      Zhang, Jian  and
      Lopyrev, Konstantin  and
      Liang, Percy",
    booktitle = "Proceedings of the 2016 Conference on Empirical Methods in Natural Language Processing",
    year = "2016",
    publisher = "Association for Computational Linguistics",
    url = "https://aclanthology.org/D16-1264/",
    pages = "2383--2392"
}

@inproceedings{triviaqa,
    title = "{T}rivia{QA}: A Large Scale Distantly Supervised Challenge Dataset for Reading Comprehension",
    author = "Joshi, Mandar  and
      Choi, Eunsol  and
      Weld, Daniel S.  and
      Zettlemoyer, Luke",
    booktitle = "Proceedings of the 55th Annual Meeting of the Association for Computational Linguistics (Volume 1: Long Papers)",
    year = "2017",
    url = "https://aclanthology.org/P17-1147/",
    pages = "1601--1611"
}

@inproceedings{blip2,
  title={Blip-2: Bootstrapping language-image pre-training with frozen image encoders and large language models},
  author={Li, Junnan and Li, Dongxu and Savarese, Silvio and Hoi, Steven},
  booktitle={International Conference on Machine Learning},
  pages={19730--19742},
  year={2023},
  organization={PMLR}
}

@inproceedings{detr,
  title={End-to-end object detection with transformers},
  author={Carion, Nicolas and Massa, Francisco and Synnaeve, Gabriel and Usunier, Nicolas and Kirillov, Alexander and Zagoruyko, Sergey},
  booktitle={European Conference on Computer Vision},
  pages={213--229},
  year={2020}
}

@inproceedings{mgh,
    title = "Negative Matters: Multi-Granularity Hard-Negative Synthesis and Anchor-Token-Aware Pooling for Enhanced Text Embeddings",
    author = "Pan, Tengyu  and
      Duan, Zhichao  and
      Li, Zhenyu  and
      Dong, Bowen  and
      Liu, Ning  and
      Li, Xiuxing  and
      Wang, Jianyong",
    booktitle = "Proceedings of the 63rd Annual Meeting of the Association for Computational Linguistics (Volume 1: Long Papers)",
    year = "2025",
    url = "https://aclanthology.org/2025.acl-long.1501/",
    pages = "31102--31118",
}

@inproceedings{sbert,
    title = "Sentence-{BERT}: Sentence Embeddings using {S}iamese {BERT}-Networks",
    author = "Reimers, Nils  and
      Gurevych, Iryna",
    booktitle = "Proceedings of the 2019 Conference on Empirical Methods in Natural Language Processing and the 9th International Joint Conference on Natural Language Processing (EMNLP-IJCNLP)",
    year = "2019",
    address = "Hong Kong, China",
    publisher = "Association for Computational Linguistics",
    pages = "3982--3992",
}

@article{anchor,
  title={Training LLMs to be Better Text Embedders through Bidirectional Reconstruction},
  author={Su, Chang and Shi, Dengliang and Huang, Siyuan and Du, Jintao and Meng, Changhua and Cheng, Yu and Wang, Weiqiang and Lin, Zhouhan},
  journal={arXiv preprint arXiv:2509.03020},
  year={2025}
}

@misc{qwen2.5,
    title = {Qwen2.5: A Party of Foundation Models},
    url = {https://qwenlm.github.io/blog/qwen2.5/},
    author = {Team, Qwen},
    month = {September},
    year = {2024}
}

@article{qwen2,
  title={Qwen2 technical report},
  author={Team, Qwen},
  journal={arXiv preprint arXiv:2407.10671},
  year={2024}
}

@article{kalm,
  title={KaLM-Embedding: Superior Training Data Brings A Stronger Embedding Model},
  author={Hu, Xinshuo and Shan, Zifei and Zhao, Xinping and Sun, Zetian and Liu, Zhenyu and Li, Dongfang and Ye, Shaolin and Wei, Xinyuan and Chen, Qian and Hu, Baotian and Wang, Haofen and Yu, Jun and Zhang, Min},
  journal={arXiv preprint arXiv:2501.01028},
  year={2025}
}

@article{zhang2025qwen3,
  title={Qwen3 Embedding: Advancing Text Embedding and Reranking Through Foundation Models},
  author={Zhang, Yanzhao and Li, Mingxin and Long, Dingkun and Zhang, Xin and Lin, Huan and Yang, Baosong and Xie, Pengjun and Yang, An and Liu, Dayiheng and Lin, Junyang and Huang, Fei and Zhou, Jingren},
  journal={arXiv preprint arXiv:2506.05176},
  year={2025}
}

@article{lee2025gemini,
  title={Gemini Embedding: Generalizable Embeddings from Gemini},
  author={Lee, Jinhyuk and Chen, Feiyang and Dua, Sahil and Cer, Daniel and Shanbhogue, Madhuri and Naim, Iftekhar and {\'A}brego, Gustavo Hern{\'a}ndez and Li, Zhe and Chen, Kaifeng and Vera, Henrique Schechter and others},
  journal={arXiv preprint arXiv:2503.07891},
  year={2025}
}

@article{dai2022promptagator,
  title={Promptagator: Few-shot dense retrieval from 8 examples},
  author={Dai, Zhuyun and Zhao, Vincent Y and Ma, Ji and Luan, Yi and Ni, Jianmo and Lu, Jing and Bakalov, Anton and Guu, Kelvin and Hall, Keith B and Chang, Ming-Wei},
  journal={arXiv preprint arXiv:2209.11755},
  year={2022}
}

@article{meng2024sfrembedding,
  title={Sfrembedding-mistral: enhance text retrieval with transfer learning},
  author={Meng, Rui and Liu, Ye and Joty, Shafiq Rayhan and Xiong, Caiming and Zhou, Yingbo and Yavuz, Semih},
  journal={Salesforce AI Research Blog},
  volume={3},
  pages={6},
  year={2024}
}

@article{team2024gemini,
  title={Gemini 1.5: Unlocking multimodal understanding across millions of tokens of context},
  author={Team, Gemini and Georgiev, Petko and Lei, Ving Ian and Burnell, Ryan and Bai, Libin and Gulati, Anmol and Tanzer, Garrett and Vincent, Damien and Pan, Zhufeng and Wang, Shibo and others},
  journal={arXiv preprint arXiv:2403.05530},
  year={2024}
}

@inproceedings{wortsman2022model,
  title={Model soups: averaging weights of multiple fine-tuned models improves accuracy without increasing inference time},
  author={Wortsman, Mitchell and Ilharco, Gabriel and Gadre, Samir Ya and Roelofs, Rebecca and Gontijo-Lopes, Raphael and Morcos, Ari S and Namkoong, Hongseok and Farhadi, Ali and Carmon, Yair and Kornblith, Simon and Schmidt, Ludwig},
  booktitle={International conference on machine learning},
  pages={23965--23998},
  year={2022},
  organization={PMLR}
}

@article{yang2025qwen3,
  title={Qwen3 Technical Report},
  author={Yang, An and Li, Anfeng and Yang, Baosong and Zhang, Beichen and Hui, Binyuan and Zheng, Bo and Yu, Bowen and Gao, Chang and Huang, Chengen and Lv, Chenxu and others},
  journal={arXiv preprint arXiv:2505.09388},
  year={2025}
}

@inproceedings{fu2025token,
  title={Token Prepending: A Training-Free Approach for Eliciting Better Sentence Embeddings from LLMs},
  author={Fu, Yuchen and Cheng, Zifeng and Jiang, Zhiwei and Wang, Zhonghui and Yin, Yafeng and Li, Zhengliang and Gu, Qing},
  booktitle={Proceedings of the 63rd Annual Meeting of the Association for Computational Linguistics (Volume 1: Long Papers)},
  pages={3168--3181},
  year={2025}
}

@inproceedings{li-li-2024-bellm,
    title = "{B}e{LLM}: Backward Dependency Enhanced Large Language Model for Sentence Embeddings",
    author = "Li, Xianming  and
      Li, Jing",
    booktitle = "Proceedings of the 2024 Conference of the North American Chapter of the Association for Computational Linguistics: Human Language Technologies (Volume 1: Long Papers)",
    year = "2024",
    url = "https://aclanthology.org/2024.naacl-long.45/",
    pages = "792--804"
}

@article{weller2025seq,
  title={Seq vs seq: An open suite of paired encoders and decoders},
  author={Weller, Orion and Ricci, Kathryn and Marone, Marc and Chaffin, Antoine and Lawrie, Dawn and Van Durme, Benjamin},
  journal={arXiv preprint arXiv:2507.11412},
  year={2025}
}

@inproceedings{compselect,
author = {Zhang, Qianchi and Zhang, Hainan and Pang, Liang and Tong, Yongxin and Zheng, Hongwei and Zheng, Zhiming},
title = {Less is More: Compact Clue Selection for Efficient Retrieval-Augmented Generation Reasoning},
year = {2026},
booktitle = {Proceedings of the ACM Web Conference 2026},
pages = {1971--1982},
}

@article{zhang2026stable,
  title={Stable-RAG: Mitigating Retrieval-Permutation-Induced Hallucinations in Retrieval-Augmented Generation},
  author={Zhang, Qianchi and Zhang, Hainan and Pang, Liang and Zheng, Hongwei and Zheng, Zhiming},
  journal={arXiv preprint arXiv:2601.02993},
  year={2026}
}

@article{cmedteb,
  title   = {CMedTEB \& CARE: Benchmarking and Enabling Efficient Chinese Medical Retrieval via Asymmetric Encoders},
  author  = {Jiang, Angqing and Chen, Jianlyu and Fang, Zhe and Wang, Yongcan and Li, Xinpeng and Ding, Keyu and Lian, Defu},
  journal = {arXiv preprint arXiv:2604.10937},
  year    = {2026},
}

\appendix

\section{Experimental Details for Training}
\label{appnendix: details_training}
\subsection{Implementation Details}
In this section, we provide additional experimental details based on Section~\ref{sec:implementation}. Following the implementation of open-source~\href{https://github.com/McGill-NLP/llm2vec}{LLM2Vec}, we employ the same fixed random seed to guarantee fair comparison and reproducibility of our results. In addition, we use the AdamW optimizer with an initial learning rate of 1e-4, and a warm-up strategy for the first 300 steps followed by linear decay over the remaining steps. To reduce GPU memory usage, all models are trained with bfloat16 quantization, gradient checkpointing, and FlashAttention-2~\citep{flashattention}. We also apply gradient accumulation to process a large batch size of 512 while sampling the same dataset within each batch. The maximum sequence length is set to the standard 512 tokens by default. 

\begin{table}[hb]
\centering
\resizebox{0.98\linewidth}{!}{
\begin{tabular}{lcc}
\toprule
\textbf{Hyperparameter} & \textbf{\texttt{1.3 B \& 1.5 B}} & \textbf{\texttt{7B}} \\
\midrule
Batch Size & 128 & 64 \\
Gradient Accumulation Steps & 4 & 8 \\
Training Steps & 2000 & 1000 \\
Maximum Steps & 4000 & 2000 \\
LoRA Rank & 64 & 64 \\
LoRA Alpha & 128 & 32 \\
LoRA for Bidirectional Encoder & \cmark & \xmark \\
\bottomrule
\end{tabular}
}
\caption{Hyperparameters used in the experiments.}
\label{tab:appendix:hyperparameters}
\end{table}

\begin{table*}[ht]
\small
\centering
\resizebox{\textwidth}{!}{
\begin{tabular}{ll}
\toprule
\textbf{Dataset} & \textbf{Instruction (s)} \\
\midrule
ELI5 & Provided a user question, retrieve the highest voted answers on Reddit ELI5 forum \\
HotpotQA & Given a multi-hop question, retrieve documents that can help answer the question \\
FEVER & Given a claim, retrieve documents that support or refute the claim \\
MIRACL & Given a question, retrieve Wikipedia passages that answer the question \\
MSMARCO Passage & Given a web search query, retrieve relevant passages that answer the query \\
MSMARCO Document & Given a web search query, retrieve relevant documents that answer the query \\
NQ & Given a question, retrieve Wikipedia passages that answer the question \\
NLI & Given a premise, retrieve a hypothesis that is entailed by the premise \\
    & Retrieve semantically similar text \\
SQuAD & Retrieve Wikipedia passages that answer the question \\
TriviaQA & Retrieve Wikipedia passages that answer the question \\
QuoraDuplicates & Given a question, retrieve questions that are semantically equivalent to the given question \\
 & Find questions that have the same meaning as the input question \\
Mr. TyDi & Given a question, retrieve Wikipedia passages that answer the question \\
DuReader & Given a Chinese search query, retrieve web passages that answer the question \\
T2Ranking & Given a Chinese search query, retrieve web passages that answer the question \\
\bottomrule
\end{tabular}}
\caption{Instructions used for public retrieval datasets.}
\label{tab:appendix:finetuning_instructions}
\end{table*}
\begin{table*}[ht]
    \centering
    \small
    \scalebox{1}{
    \begin{tabular}{ll}
    \toprule
    Category & Dataset \\ \toprule
    \multicolumn{1}{l}{\multirow{1}{*}{Retrieval (6)}} &  SciFact, NFCorpus, FiQA2018, SCIDOCS, TRECCOVID, Touche2020\\
    \midrule
    
    \multicolumn{1}{l}{\multirow{1}{*}{Reranking (3)}} & AskUbuntuDupQuestions, SciDocsRR, StackOverflowDupQuestions\\
    \midrule

    \multicolumn{1}{l}{\multirow{2}{*}{Clustering (5)}} & BiorxivClusteringS2S, BiorxivClusteringP2P, MedrxivClusteringS2S,\\
    & MedrxivClusteringP2P, TwentyNewsgroupsClustering\\
    \midrule

    \multicolumn{1}{l}{\multirow{1}{*}{Pair Classification (2)}} & SprintDuplicateQuestions, TwitterURLCorpus\\ 
    \midrule

    \multicolumn{1}{l}{\multirow{2}{*}{Classification (6)}} & AmazonReviewsClassification, Banking77Classification, EmotionClassification,\\
    & MTOPDomainClassification, TweetSentimentExtractionClassification, ImdbClassification \\
    \midrule

    \multicolumn{1}{l}{\multirow{1}{*}{STS (7)}} & BIOSSES, STS12, STS13, STS14, STS15, STS16, STSBenchmark\\
    \midrule
    
    \multicolumn{1}{l}{\multirow{1}{*}{SummEval (1)}} & SummEval\\
    \midrule
    
    \multicolumn{1}{l}{\multirow{1}{*}{Overall}} & 30 datasets\\
    \bottomrule
    \end{tabular}}
    \caption{Composition of the MTEB-MINI benchmark.}
    \label{tab:mteb-mini}
\end{table*}
Table~\ref{tab:appendix:hyperparameters} presents the hyperparameters that vary across different base models. We set the LoRA~\cite{hu2022lora} rank to 64 and LoRA alpha to 32 for 7B models following~\citet{bgeicl}, while a large LoRA alpha of 128 is used for smaller models. For training steps, 7B models are trained for 1000 steps as in~\citet{llm2vec, mgh}, whereas smaller models require 2000 steps. The maximum steps for the linear scheduler are set to twice the training steps. In terms of computational cost, training a 7B model typically requires about 33 A100 80GB GPU hours, while the 1.3B and 1.5B models require over 20 GPU hours.

As we adopt instruction-tuned versions of Qwen2.5-1.5B, LLaMA-2-7B, and Mistral-7B, special tokens are added to the input text according to the official instruction prompt templates for each model. Specifically, \texttt{[INST]} and \texttt{[/INST]} are used for LLaMA-2-7B and Mistral-7B, while \texttt{<|im\_start|>user\textbackslash n} and \texttt{<|im\_end|>} are added for Qwen2.5-1.5B. Moreover, to insert the Contextual token into LLM's input sequence, we first add a placeholder to the input text: \texttt{<s>} for S-LLaMA-1.3B, LLaMA-2-7B, and Mistral-7B, and \texttt{<|end\_of\_text|>} for Qwen2.5-1.5B. After tokenizing the modified input and obtaining its word embeddings from LLM's embedding layer, we replace the placeholder with the Contextual token.  

\subsection{HuggingFace Models}
All base models and bidirectional encoders employed in this work are obtained from the HuggingFace platform: 
\paragraph{Base Models}
\begin{itemize}[nosep, left=0pt]
\item S-LLaMA-1.3B~\cite{sheared}:~\href{https://huggingface.co/princeton-nlp/Sheared-LLaMA-1.3B}{princeton-nlp/Sheared-LLaMA-1.3B}
\item Qwen2.5-1.5B~\cite{qwen2, qwen2.5}:~\href{https://huggingface.co/Qwen/Qwen2.5-1.5B-Instruct}{Qwen/Qwen2.5-1.5B-Instruct}
\item LLaMA-2-7B~\cite{llama2}:~\href{https://huggingface.co/meta-llama/Llama-2-7b-chat-hf}{meta-llama/Llama-2-7b-chat-hf}
\item Mistral-7B~\cite{mistral}:~\href{https://huggingface.co/mistralai/Mistral-7B-Instruct-v0.2}{mistralai/Mistral-7B-Instruct-v0.2}
\end{itemize} 
\paragraph{Bidirectional Encoders}
\begin{itemize}[nosep, left=0pt]
\item GTE-small~\cite{gte}:~\href{https://huggingface.co/thenlper/gte-small}{thenlper/gte-small}
\item E5-small-v2~\cite{e5}:~\href{https://huggingface.co/intfloat/e5-small-v2}{intfloat/e5-small-v2}
\item E5-base-v2~\cite{e5}:~\href{https://huggingface.co/intfloat/e5-base-v2}{intfloat/e5-base-v2}
\item E5-large-v2~\cite{e5}:~\href{https://huggingface.co/intfloat/e5-large-v2}{intfloat/e5-large-v2}
\end{itemize} 

\subsection{Public Retrieval Datasets and Instructions}
\label{appendix:training_data}
The collection of publicly available retrieval datasets used for training is curated by~\citet{echo} and is distributed under the \href{https://github.com/jakespringer/echo-embeddings/blob/master/LICENSE}{Apache License 2.0}. It includes the following datasets: ELI5 (sample ratio 0.1)~\citep{eli5}, HotpotQA~\citep{hotpotqa}, FEVER~\citep{fever}, MIRACL~\citep{miracl}, MS-MARCO passage ranking (sample ratio 0.5) and document ranking (sample ratio 0.2)~\citep{marco}, NQ~\citep{karpukhin2020dense}, NLI~\citep{simcse}, SQuAD~\citep{squad}, TriviaQA~\citep{triviaqa}, Quora Duplicate Questions (sample ratio 0.1)~\citep{quora}, Mr. TyDi~\citep{MrTyDi}, DuReader~\citep{dureader}, and T2Ranking (sample ratio 0.5)~\citep{t2ranking}. 

Table~\ref{tab:appendix:finetuning_instructions} lists the instructions used for each dataset, which are manually written by~\citet{e5mistral}. Notably, in the query-passage pairs of retrieval datasets, task-specific instructions are appended only to the queries, without modifying the passages.

\section{Experimental Details for Evaluation}
\label{appnendix: details_evaluation}
\subsection{MTEB-MINI Details}
\label{appendix:mteb-mini}
Considering the substantial computational resources required for full evaluation on MTEB, we follow~\citet{echo, mgh} and select a subset of MTEB for ablation and analysis. While prior studies~\cite{llm2vec, anchor} utilize only a few datasets, our preliminary experiments suggest that evaluation on a limited subset may introduce significant bias and fail to effectively reflect the overall trends of the full MTEB. To this end, as shown in Table~\ref{tab:mteb-mini}, we empirically introduce the MTEB-MINI by selecting 30 representative datasets spanning all task categories in MTEB.

\subsection{Evaluation Metrics}
The task categories of MTEB include Retrieval (Retr.), Reranking (Rerank.), Clustering (Clust.), Pair Classification (PairClass.), Classification (Class.), Semantic Textual Similarity (STS), and Summarization (Summ.). For these tasks, the main evaluation metrics are nDCG@10, MAP, V-measure (V-meas.), average precision (AP), accuracy (Acc.), and Spearman correlation (Spear., both for STS and Summ.), respectively.

\subsection{Instructions for MTEB Evaluation}
To enable a fair comparison with prior leading embedding methods~\citep{e5mistral, echo, llm2vec, nvembed, bgeicl,mgh}, we use the same instruction prompts for evaluation on both MTEB and MTEB-MINI. The instructions applied to each dataset are listed in Table~\ref{tab:appendix:mteb_instructions}.

\begin{figure*}[t]
    \centering
    \includegraphics[width=0.95\textwidth]{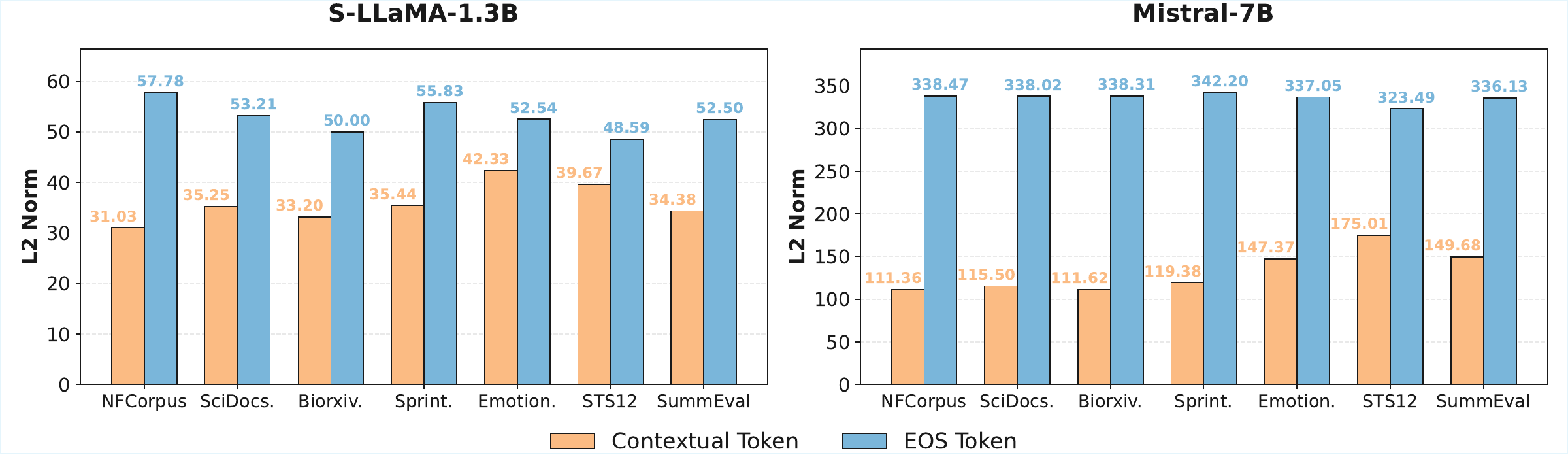}
    \caption{L2 norms of Contextual and EOS tokens on selected MTEB subsets for two base models: S-LLaMA-1.3B and Mistral-7B. The evaluated datasets span seven tasks, including NFCorpus, SciDocsRR (SciDocs.), BiorxivClusteringP2P (Biorxiv.), SprintDuplicateQuestions (Sprint.), EmotionClassification (Emotion.), STS12, and SummEval.}
    \label{fig:l2_norm}
\end{figure*}

\section{Additional Results}

\subsection{The L2 Norms of Contextual and EOS Tokens}
To examine the respective contributions of Contextual and EOS tokens to the final text embedding, we compare their L2 norms on selected MTEB datasets. As depicted in Figure~\ref{fig:l2_norm}, we observe that the EOS token consistently shows higher L2 norms across various task categories, indicating its greater influence on the concatenated representation.

\begin{table}[h]
\centering
\scalebox{0.73}{
\begin{tabular}{lccc}
\toprule
\textbf{Method} & \textbf{S-LLaMA-1.3B} & \textbf{Qwen2.5-1.5B} & \textbf{Mistral-7B}\\
\midrule
w/ Bi-LoRA & \textbf{62.83} & \textbf{64.05} & 65.64 \\
w/o Bi-LoRA & 62.57 & 63.82 & \textbf{65.85} \\
\bottomrule 
\end{tabular}
}
\caption{Performance comparison on MTEB-MINI (30 datasets) between Causal2Vec with and without applying LoRA to the bidirectional encoder (\texttt{Bi-LoRA}), using three base models: S-LLaMA-1.3B, Qwen2.5-1.5B, and Mistral-7B.}
\label{tab:detail_bi_lora}
\end{table}

\begin{table}[t]
\centering
\scalebox{0.85}{
\begin{tabular}{l cc}
\toprule
\textbf{Method} & \textbf{Retr.}  & \textbf{MTEB-MINI} \\
\midrule
e5-mistral-7b-instruct  &  49.10  & 65.75 \\
\rowcolor{blue!10}Causal2Vec & 51.22 ({\textcolor[rgb]{0.86,0.20,0.18}{+2.12}})  &66.59 ({\textcolor[rgb]{0.86,0.20,0.18}{+0.84}})  \\
\bottomrule 
\end{tabular}
}
\caption{Average scores on the retrieval task (6 datasets) and MTEB-MINI (30 datasets) for \href{https://huggingface.co/intfloat/e5-mistral-7b-instruct}{e5-mistral-7b-instruct} and Causal2Vec$_\text{e5-mistral-7b-instruct}$.}
\label{tab:existing_embedding}
\end{table}

\subsection{Impact of Freezing the Bidirectional Encoder}
\label{appendix:freeze_bert}
Since the BERT-style bidirectional encoder (E5-base-v2) we use is specifically trained for embedding tasks, this section investigates whether it should be frozen during fine-tuning. As shown in Table~\ref{tab:detail_bi_lora}, we observe that fine-tuning the bidirectional encoder with LoRA leads to an average score improvement of $0.26$ and $0.23$ on the MTEB-MINI for S-LLaMA-1.3B and Qwen2.5-1.5B, respectively, but degrades Mistral-7B's performance by $0.21$ points. We attribute this to two potential effects of making the bidirectional encoder trainable: (1) it may cause catastrophic forgetting in the bidirectional encoder, and (2) it may help the LLM better interpret the Contextual token through joint fine-tuning. Large-scale LLMs are more susceptible to the former, as they already have sufficient capacity to comprehend newly added tokens during fine-tuning~\citep{llava, blip2}. This suggests that whether the introduced bidirectional encoder should remain frozen may depend on the scale of the underlying LLM.

\subsection{Effectiveness on Existing Embedding Models}
In this section, we examine the effectiveness of Causal2Vec on existing high-performing embedding models rather than on general-purpose LLMs. As shown in Table~\ref{tab:existing_embedding}, Causal2Vec remains effective when applied to e5-mistral-7b-instruct~\cite{e5mistral}, a powerful embedding model that uses causal attention with last-token pooling and is fine-tuned on extensive synthetic data. This demonstrates that our method indeed mitigates the inherent limitations of causal attention in representation learning. Moreover, we observe larger improvements on retrieval tasks involving long text inputs, further confirming the effectiveness of our proposed representation method.

\subsection{Different Representation Methods}
\label{appendix:representation_type}
In this section, we explore various representation methods tailored to our embedding framework through the following experimental settings (Note: all experiments incorporate the Contextual token into the LLM's input sequence):
\begin{enumerate}[itemsep=0pt]
\item \textbf{Default:} Causal2Vec generates the text embedding by concatenating LLM's output hidden states of the Contextual and EOS tokens.
\item \textbf{Concat-bi:} Concatenate the LLM's output EOS token with the output of the bidirectional encoder processed by an MLP layer.
\item \textbf{Average:} Take the average of the LLM’s output hidden states corresponding to the Contextual and EOS tokens.
\item \textbf{Last-token pooling:} Use only the final hidden states of the last EOS token as text embedding.
\end{enumerate}
\begin{table}[h]
\centering
\scalebox{0.9}{
\begin{tabular}{l cc}
\toprule
\textbf{Method} & \textbf{S-LLaMA-1.3B} &\textbf{Mistral-7B}\\
\midrule
\rowcolor{blue!10}Causal2Vec  &  \textbf{62.83} & \textbf{65.85}\\
\quad w/ Concat-bi & 62.44 & 65.59\\
\quad w/ Average & 62.67 & 65.55\\
\quad w/ Last-token & 62.19 & 65.44\\
\bottomrule 
\end{tabular}
}
\caption{Performance comparison on MTEB-MINI (30 datasets) using different representation method with two base models: S-LLaMA-1.3B and Mistral-7B.}
\label{tab:pooling_type}
\end{table}

\begin{table}[h]
\centering
\scalebox{0.88}{
\begin{tabular}{l cccc}
\toprule
\textbf{Method} & \textbf{Dim.} &\textbf{ArguAna} &\textbf{SciFace} &\textbf{NFCorpus}\\
\midrule
\rowcolor{blue!10}Default  & 8192 &  324.08s &243.10s &173.81s\\
Average & 4096 & 322.70s &242.02s &173.27s \\
\bottomrule 
\end{tabular}
}
\caption{End-to-end completion time (in seconds) for Causal2Vec$_\text{mistral-7b}$ across retrieval datasets with different output embedding dimensions.}
\label{tab:efficiency_concat_avg}
\end{table}

Table~\ref{tab:pooling_type} presents the comparison results on MTEB-MINI, from which we draw the following observations: (1) The representation methods that utilize both the EOS and Contextual tokens consistently outperform last-token pooling for S-LLaMA-1.3B and Mistral-7B. This suggests that incorporating an additional context-aware token with the EOS token leads to richer semantic information while reducing the model’s reliance on the single EOS token alone. (2) We observe further performance improvements when the concatenated Contextual token is derived from the LLM, rather than from the bidirectional encoder followed by an MLP layer. We speculate that this helps the LLM better capture the semantic content encoded in the Contextual token. (3) Among the strategies utilizing the last hidden states of both EOS and Contextual tokens, concatenation yields substantially better text representations than averaging.

We further compare the overall completion time across retrieval datasets using the concatenation and averaging representation methods. As shown in Table~\ref{tab:efficiency_concat_avg}, although Causal2Vec doubles the output embedding dimension under concatenation, the additional runtime overhead is negligible, suggesting that LLM inference remains the dominant computational bottleneck in practice.

\begin{table}[t]
\centering
\scalebox{0.9}{
\begin{tabular}{c cc}
\toprule
\textbf{Method} & \textbf{S-LLaMA-1.3B}  & \textbf{Mistral-7B} \\
\midrule
before instruction  &  62.69  &65.55 \\
\rowcolor{blue!10}after instruction & \textbf{62.83}  &\textbf{65.85}  \\
\bottomrule 
\end{tabular}
}
\caption{Average MTEB-MINI (30 datasets) score for placing the Contextual token before and after the task-specific instruction in Causal2Vec.}
\label{tab:order_of_contextual}
\end{table}

\subsection{The Position of the Contextual Token.}
We investigate whether the position of the Contextual token affects embedding performance by comparing two placement settings: before vs. after the instruction. As shown in Table~\ref{tab:order_of_contextual}, "\texttt{after instruction}" consistently yields better results. We speculate that positioning the Contextual token before the instruction tokens may hinder the LLM’s ability to accurately interpret and follow task-specific prompts.

\subsection{Impact of Different Attention Mechanisms}
\label{appendix:different_attention}
This section verifies the impact of different attention mechanisms in our proposed Causal2Vec. As illustrated in Table~\ref{tab:attention}, shifting from causal to bidirectional attention consistently leads to lower performance of on MTEB-MINI, even for the Mistral-7B backbone, which generally benefits from attention modifications in most bidirectional attention-based embedding methods~\cite{grit,llm2vec,nvembed,mgh}. We hypothesize that the attention mismatch between pre-training and contrastive learning compromises LLM's ability to extract the well-learned semantic information. This finding further highlights that altering the original attention mechanism of LLMs may be suboptimal for text embedding tasks.
\begin{table}[t]
\centering
\scalebox{0.73}{
\begin{tabular}{lccc}
\toprule
\textbf{Attention} & \textbf{S-LLaMA-1.3B} & \textbf{LLaMA-2-7B} & \textbf{Mistral-7B}\\
\midrule
\rowcolor{blue!10}Causal & \textbf{62.83} & \textbf{64.62} & \textbf{65.85} \\
Bidirectional & 62.52 & 64.30 & 65.77 \\
\bottomrule 
\end{tabular}
}
\caption{Performance comparison of Causal2Vec using bidirectional and causal attention mechanisms on MTEB-MINI (30 datasets), with three base models: S-LLaMA-1.3B, LLaMA-2-7B, and Mistral-7B.}
\label{tab:attention}
\end{table}

\begin{figure}[h]
    \centering
    \includegraphics[scale=0.42]{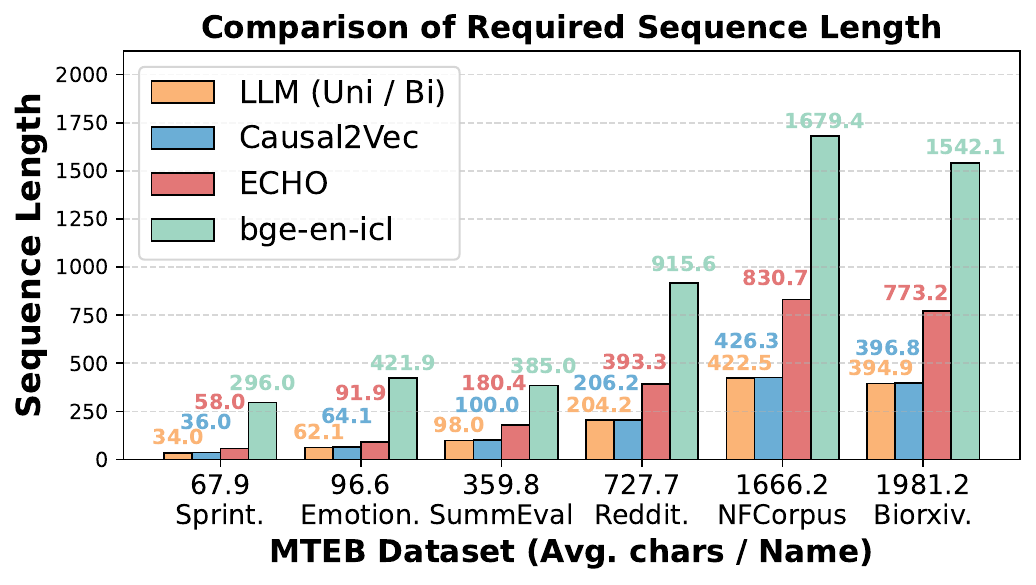}
    \caption{Average required sequence length per sample for various Mistral-7B-based methods on selected MTEB subsets. \texttt{LLM (Uni/Bi)} denotes the standard Mistral-7B with causal or bidirectional attention. For the asymmetric dataset NFCorpus, we report the results per query-passage pair. Note: Echo~\cite{echo} repeats only the input text, excluding the instruction. For bge-en-icl~\cite{bgeicl}, in-context examples are taken from the official \href{https://github.com/FlagOpen/FlagEmbedding/tree/master/FlagEmbedding/evaluation/mteb/examples}{repository}.}
    \label{fig:length_comparison}
\end{figure}

\subsection{Comparison on Required Sequence Length}
\label{appnendix: length_comparison}
Figure~\ref{fig:length_comparison} presents the approximate average sequence lengths required by different models on selected MTEB subsets. Specifically, ECHO~\cite{echo} repeats the input, while bge-en-icl~\cite{bgeicl} incorporates several task-related in-context examples, both of which significantly increase the required sequence length for LLMs. In contrast, our method requires only a single additional Contextual token compared to standard LLMs. Notably, when using \texttt{<s>} as a placeholder for the Contextual token in Mistral-7B, tokenization produces not only the \texttt{<s>} token itself but also an additional separator token. As a result, Causal2Vec reduces the required sequence length by up to \textbf{85\%} (Sprint.: 34.0 vs. 269.0; Emotion.: 62.1 vs. 421.9) compared to the state-of-the-art unidirectional method bge-en-icl.

\subsection{Full MTEB Results}
We present detailed results on all 56 MTEB datasets for the proposed Causal2Vec in Table~\ref{tab:appendix:full-mteb-results-1} and Table~\ref{tab:appendix:full-mteb-results-2}, including four base models: S-LLaMA-1.3B, Qwen2.5-1.5B, LLaMA-2-7B, and Mistral-7B.
\label{appendix:full_mteb}

\section{Related Works}
Many industry-developed embedding models, such as KaLM-Embedding~\cite{kalm}, Gemini Embedding~\cite{lee2025gemini}, and Qwen3 Embedding~\cite{zhang2025qwen3}, have achieved remarkably outstanding performance on MTEB. However, these approaches heavily rely on extensive proprietary synthetic data for training as well as various engineering optimizations, making direct comparison with academic research unfair~\cite{bgeicl,anchor}. Therefore, we do not include these models in our comparisons, we instead briefly introduce them to highlight their contributions.

KaLM-Embedding~\cite{kalm} is a superior embedding model that aims to improve the training data quality and distills knowledge from LLMs into text embeddings. Specifically, it is trained on a carefully curated corpus that includes more than 20 data categories for pre-training and 70 categories for fine-tuning. In addition, KaLM-Embedding applies several key techniques~\cite{dai2022promptagator,meng2024sfrembedding,e5mistral} to further enhance and clean the data, resulting in a substantially larger and higher-quality training dataset.

Gemini Embedding~\cite{lee2025gemini} is built upon the powerful Gemini LLM~\cite{team2024gemini} and trained on a wide range of embedding tasks. It leverages Gemini to guide the construction of a diverse and high-quality training dataset. Furthermore, the final embedding model is obtained by combining several fine-tuned checkpoints using an effective parameter-averaging technique~\cite{wortsman2022model}.

Qwen3 Embedding series~\cite{zhang2025qwen3} are built upon the Qwen3 foundation models~\cite{yang2025qwen3} and exhibit strong performance in text embedding and reranking. The training pipeline of Qwen3 Embedding combines large-scale unsupervised pre-training with supervised fine-tuning on extensive training datasets. In particular, the high-quality and diverse training data covering multiple domains and languages is synthesized by the Qwen3 LLMs.

\section{Ethical Considerations}
Our proposed Causal2Vec can be applied to a wide range of real-world applications, including information retrieval and LLM-based retrieval-augmented generation systems. However, LLMs are known to suffer from biases and hallucinations, which could potentially lead to negative social impacts.

\begin{table*}[h]
\centering   
\resizebox{\textwidth}{!}{
\begin{tabular}{ll}
\toprule
\textbf{Dataset} & \textbf{Instruction Template} \\
\midrule
AmazonCounterfactualClassification & \begin{tabular}[c]{@{}l@{}}Classify a given Amazon customer review text as either counterfactual \\ or not-counterfactual.\end{tabular} \\
AmazonPolarityClassification & Classify Amazon reviews into positive or negative sentiment  \\
AmazonReviewsClassification & Classify the given Amazon review into its appropriate rating category  \\
Banking77Classification & Given a online banking query, find the corresponding intents  \\
EmotionClassification &  \begin{tabular}[c]{@{}l@{}}Classify the emotion expressed in the given Twitter message into one of the six \\ emotions: anger, fear, joy, love, sadness, and surprise.\end{tabular}   \\
ImdbClassification & Classify the sentiment expressed in the given movie review text from the IMDB dataset.\\
MassiveIntentClassification & Given a user utterance as query, find the user intents  \\
MassiveScenarioClassification & Given a user utterance as query, find the user scenarios  \\
MTOPDomainClassification & Classify the intent domain of the given utterance in task-oriented conversation  \\
MTOPIntentClassification & Classify the intent of the given utterance in task-oriented conversation  \\
ToxicConversationsClassif. & Classify the given comments as either toxic or not toxic  \\
TweetSentimentClassification & Classify the sentiment of a given tweet as either positive, negative, or neutral  \\
ArxivClusteringP2P & Identify the main and secondary category of Arxiv papers based on the titles and abstracts. \\
ArxivClusteringS2S & Identify the main and secondary category of Arxiv papers based on the titles  \\
BiorxivClusteringP2P & Identify the main category of Biorxiv papers based on the titles and abstracts  \\
BiorxivClusteringS2S & Identify the main category of Biorxiv papers based on the titles  \\
MedrxivClusteringP2P & Identify the main category of Medrxiv papers based on the titles and abstracts  \\
MedrxivClusteringS2S & Identify the main category of Medrxiv papers based on the titles  \\
RedditClustering & Identify the topic or theme of Reddit posts based on the titles  \\
RedditClusteringP2P & Identify the topic or theme of Reddit posts based on the titles and posts  \\
StackExchangeClustering & Identify the topic or theme of StackExchange posts based on the titles  \\
StackExchangeClusteringP2P & Identify the topic or theme of StackExchange posts based on the given paragraphs  \\
TwentyNewsgroupsClustering & Identify the topic or theme of the given news articles  \\
SprintDuplicateQuestions & Retrieve duplicate questions from Sprint forum  \\
TwitterSemEval2015 & Retrieve tweets that are semantically similar to the given tweet  \\
TwitterURLCorpus & Retrieve tweets that are semantically similar to the given tweet  \\
AskUbuntuDupQuestions & Retrieve duplicate questions from AskUbuntu forum  \\
MindSmallReranking & Retrieve relevant news articles based on user browsing history  \\
SciDocsRR & Given a title of a scientific paper, retrieve the titles of other relevant papers  \\
StackOverflowDupQuestions & Retrieve duplicate questions from StackOverflow forum  \\
ArguAna & Given a claim, find documents that refute the claim  \\
ClimateFEVER & Given a claim about climate change, retrieve documents that support or refute the claim. \\
CQADupstackRetrieval &  \begin{tabular}[c]{@{}l@{}}Given a question, retrieve detailed question descriptions from Stackexchange that are \\ duplicates to the given question.\end{tabular} \\
DBPedia & Given a query, retrieve relevant entity descriptions from DBPedia  \\
FEVER & Given a claim, retrieve documents that support or refute the claim  \\
FiQA2018 & Given a financial question, retrieve user replies that best answer the question  \\
HotpotQA & Given a multi-hop question, retrieve documents that can help answer the question  \\
MSMARCO & Given a web search query, retrieve relevant passages that answer the query  \\
NFCorpus & Given a question, retrieve relevant documents that best answer the question  \\
NQ & Given a question, retrieve Wikipedia passages that answer the question  \\
QuoraRetrieval & Given a question, retrieve questions that are semantically equivalent to the given question. \\
SCIDOCS & Given a scientific paper title, retrieve paper abstracts that are cited by the given paper  \\
SciFact & Given a scientific claim, retrieve documents that support or refute the claim  \\
Touche2020 & Given a question, retrieve detailed and persuasive arguments that answer the question  \\
TRECCOVID & Given a query on COVID-19, retrieve documents that answer the query  \\
STS* & Retrieve semantically similar text.  \\
SummEval & Given a news summary, retrieve other semantically similar summaries  \\
\bottomrule
\end{tabular}}
\caption{Instructions used for evaluation on the MTEB. ``STS*'' denotes that the corresponding instruction is applied to all STS datasets.}
\label{tab:appendix:mteb_instructions}
\end{table*}
\begin{table*}[t]
    \centering
    \small
    \begin{tabular}{l|ccc}
    \toprule
    \textbf{Dataset} & \textbf{\texttt{S-LLaMA-1.3B}}  & \textbf{\texttt{Qwen2.5-1.5B}} & \textbf{\texttt{LLaMA-2-7B}} \\
    \midrule
    AmazonCounterfactualClassification & 74.49 & 73.04 & 76.79 \\
    AmazonPolarityClassification & 92.75 & 94.23 & 94.80 \\
    AmazonReviewsClassification & 46.48 & 46.61 & 51.75 \\
    ArguAna & 54.73 & 58.59 & 57.35 \\
    ArxivClusteringP2P & 46.25 & 49.16 & 48.37 \\
    ArxivClusteringS2S & 39.52 & 44.97 & 42.88 \\
    AskUbuntuDupQuestions & 61.59 & 64.18 & 63.54 \\
    BIOSSES & 83.96 & 85.94 & 84.06 \\
    Banking77Classification & 85.96 & 86.74 & 88.14 \\
    BiorxivClusteringP2P & 38.13 & 38.71 & 39.05 \\
    BiorxivClusteringS2S & 35.13 & 36.80 & 36.42 \\
    CQADupstackRetrieval & 39.53 & 43.77 & 43.42 \\
    ClimateFEVER & 32.62 & 34.70 & 32.46 \\
    DBPedia & 44.85 & 45.88 & 49.85 \\
    EmotionClassification & 46.82 & 48.52 & 49.74 \\
    FEVER & 88.11 & 90.02 & 90.53 \\
    FiQA2018 & 44.52 & 48.41 & 51.29 \\
    HotpotQA & 67.13 & 68.43 & 71.45 \\
    ImdbClassification & 83.76 & 83.71 & 88.33 \\
    MSMARCO & 40.88 & 41.41 & 41.22 \\
    MTOPDomainClassification & 94.05 & 94.60 & 95.53 \\
    MTOPIntentClassification & 73.45 & 78.37 & 82.38 \\
    MassiveIntentClassification & 73.36 & 75.78 & 77.63 \\
    MassiveScenarioClassification & 77.58 & 78.91 & 79.88 \\
    MedrxivClusteringP2P & 33.38 & 34.31 & 33.13 \\
    MedrxivClusteringS2S & 31.58 & 32.64 & 32.14 \\
    MindSmallReranking & 32.71 & 32.47 & 32.46 \\
    NFCorpus & 37.43 & 39.59 & 40.21 \\
    NQ & 58.02 & 59.51 & 64.10 \\
    QuoraRetrieval & 88.91 & 89.39 & 88.80 \\
    RedditClustering & 56.91 & 57.66 & 63.07 \\
    RedditClusteringP2P & 61.26 & 60.78 & 64.31 \\
    SCIDOCS & 19.56 & 20.64 & 21.28 \\
    SICK-R & 81.99 & 82.70 & 82.78 \\
    STS12 & 77.04 & 79.63 & 78.77 \\
    STS13 & 87.47 & 88.46 & 88.89 \\
    STS14 & 83.21 & 84.27 & 85.29 \\
    STS15 & 88.93 & 89.64 & 89.86 \\
    STS16 & 86.83 & 87.43 & 87.72 \\
    STS17 & 91.13 & 91.53 & 92.19 \\
    STS22 & 69.21 & 68.16 & 70.67 \\
    STSBenchmark & 87.85 & 88.53 & 88.77 \\
    SciDocsRR & 81.68 & 83.74 & 84.11 \\
    SciFact & 73.04 & 74.67 & 75.77 \\
    SprintDuplicateQuestions & 96.26 & 96.57 & 97.00 \\
    StackExchangeClustering & 64.27 & 67.86 & 69.14 \\
    StackExchangeClusteringP2P & 32.41 & 35.09 & 36.74 \\
    StackOverflowDupQuestions & 50.16 & 52.70 & 52.61 \\
    SummEval & 31.45 & 30.99 & 31.09 \\
    TRECCOVID & 76.24 & 76.97 & 78.65 \\
    Touche2020 & 24.74 & 26.49 & 22.83 \\
    ToxicConversationsClassification & 65.03 & 62.07 & 65.02 \\
    TweetSentimentExtractionClassification & 61.53 & 61.61 & 61.44 \\
    TwentyNewsgroupsClustering & 49.01 & 51.29 & 54.33 \\
    TwitterSemEval2015 & 75.24 & 76.54 & 79.78 \\
    TwitterURLCorpus & 87.03 & 87.05 & 86.77 \\
    \midrule
    \textbf{MTEB Average (56)} & 62.63 & 63.97 & 64.94 \\
    \bottomrule
    \end{tabular}
    \caption{Results of Causal2Vec on all 56 MTEB datasets across three base models: S-LLaMA-1.3B, Qwen2.5-1.5B, and LLaMA-2-7B.}
    \label{tab:appendix:full-mteb-results-1}
\end{table*}

\begin{table*}[t]
    \centering
    \small
    \begin{tabular}{l|cc}
    \toprule
    \textbf{Dataset} & \textbf{\texttt{Mistral-7B}} & \textbf{\texttt{Mistral-7B (w/ ICL)}}\\
    \midrule
    AmazonCounterfactualClassification & 76.22 & 75.99\\
    AmazonPolarityClassification & 95.02 & 95.80\\
    AmazonReviewsClassification & 51.40 & 53.78\\
    ArguAna & 57.55 & 59.11\\
    ArxivClusteringP2P & 48.99 & 50.64\\
    ArxivClusteringS2S & 45.51 & 47.09\\
    AskUbuntuDupQuestions & 65.96 & 65.71\\
    BIOSSES & 86.42 &87.28\\
    Banking77Classification & 88.62 &88.85\\
    BiorxivClusteringP2P & 39.24 &40.82\\
    BiorxivClusteringS2S & 38.32 &39.09\\
    CQADupstackRetrieval & 45.59 &46.82\\
    ClimateFEVER & 35.55 &34.12\\
    DBPedia & 51.65 &52.09\\
    EmotionClassification & 50.56 &51.40\\
    FEVER & 91.53 &91.68\\
    FiQA2018 & 54.96 &56.03\\
    HotpotQA & 74.25 &73.11\\
    ImdbClassification & 91.24 &92.45\\
    MSMARCO & 42.22 &42.83\\
    MTOPDomainClassification & 95.79 &96.36\\
    MTOPIntentClassification & 83.12 &86.23\\
    MassiveIntentClassification & 78.14 &79.53\\
    MassiveScenarioClassification & 81.27 &82.40\\
    MedrxivClusteringP2P & 34.33 &36.32\\
    MedrxivClusteringS2S & 34.28 &34.73\\
    MindSmallReranking & 32.32 &32.31\\
    NFCorpus & 41.63 &41.41\\
    NQ & 66.65 &66.50\\
    QuoraRetrieval & 89.35 &89.24\\
    RedditClustering & 64.73 &64.99\\
    RedditClusteringP2P & 66.43 &68.21\\
    SCIDOCS & 22.40 &22.76\\
    SICK-R & 83.49 &83.33\\
    STS12 & 79.37 &79.71\\
    STS13 & 88.69 &89.75\\
    STS14 & 85.43 &85.80\\
    STS15 & 90.76 &90.92\\
    STS16 & 88.26 &88.38\\
    STS17 & 92.47 &92.31\\
    STS22 & 69.44 &69.20\\
    STSBenchmark & 89.47 &89.96\\
    SciDocsRR & 84.40 &84.61\\
    SciFact & 77.52 &77.92\\
    SprintDuplicateQuestions & 96.70 &97.16\\
    StackExchangeClustering & 72.23 &76.40\\
    StackExchangeClusteringP2P & 37.73 &40.63\\
    StackOverflowDupQuestions & 55.16 &54.81\\
    SummEval & 30.57 &30.82\\
    TRECCOVID & 83.48 &83.09\\
    Touche2020 & 24.86 &25.51\\
    ToxicConversationsClassification & 63.05 &65.14\\
    TweetSentimentExtractionClassification & 62.52 &62.46\\
    TwentyNewsgroupsClustering & 56.01 &59.62\\
    TwitterSemEval2015 & 81.35 &82.91\\
    TwitterURLCorpus & 87.23 &87.50\\
    \midrule
    \textbf{MTEB Average (56)} & 66.10 &66.85\\
    \bottomrule
    \end{tabular}
    \caption{Results of Causal2Vec on all 56 MTEB datasets for Mistral-7B and Mistral-7B (w/ ICL).}
    \label{tab:appendix:full-mteb-results-2}
\end{table*}

\end{document}